\documentclass[a4paper,fleqn]{cas-dc}

\usepackage[authoryear]{natbib}

\usepackage{booktabs}
\usepackage{multirow}
\usepackage{xcolor}
\usepackage{colortbl}
\usepackage{adjustbox}
\definecolor{bestred}{RGB}{180,0,0}

\def\tsc#1{\csdef{#1}{\textsc{\lowercase{#1}}\xspace}}
\tsc{WGM}
\tsc{QE}
\begin{document}
\let\WriteBookmarks\relax
\def\floatpagepagefraction{1}
\def\textpagefraction{.001}

\shorttitle{}    

\shortauthors{}  

\title [mode = title]{A Dynamic Scene Interaction Reasoning Framework for Scene-level Lane-Change Intention and Trajectory Prediction of Multiple Interacting Vehicles}  

\tnotemark[1] 

\tnotetext[1]{} 



\author[1]{Joshua Kofi Asamoah}[orcid=0009-0002-3258-0479]

% \cormark[1]

\fnmark[1]

\ead{joshua.asamoah@ndsu.edu}

\ead[url]{https://joshuakasamoah.github.io/}

\credit{Conceptualization of this study, Methodology, Writing - Original draft preparation, Writing - Review}

\affiliation[1]{organization={North Dakota State University},
            addressline={1410 14th Avenue Offerdahl North Building, CIE 201}, 
            city={Fargo},
%          citysep={}, % Uncomment if no comma needed between city and postcode
            postcode={58102}, 
            state={North Dakota},
            country={United States}}

\author[1]{Blessing Agyei Kyem}[orcid=0009-0006-6360-6386]

% Footnote of the second author
\fnmark[2]

% Email id of the second author
\ead{blessing.agyeikyem@ndsu.edu}

\ead[url]{https://blessing-agyei-kyem.github.io/}

\credit{Data Curation, Writing - Review and Editing, Validation}

\author[1]{Eugene Denteh}[orcid=0009-0001-7597-6732]

% Footnote of the second author
\fnmark[3]

% Email id of the second author
\ead{eugene.denteh@ndsu.edu}

% \ead[url]{https://blessing-agyei-kyem.github.io/}

\credit{Data Curation, Methodology, Writing - Review and Editing, Validation}

% \affiliation[1]{organization={North Dakota State University},
%             city={Fargo},
% %          citysep={}, % Uncomment if no comma needed between city and postcode
%             postcode={58102}, 
%             state={North Dakota},
%             country={United States}}

\author[1]{Armstrong Aboah}[orcid=0000-0002-1605-1545]
\cormark[1]
\ead{armstrong.aboah@ndsu.edu}
\ead[url]{https://aboaharmstrong.vercel.app/}
\credit{Conceptualization of this study, Methodology, Writing - Original draft preparation, Writing - Review and Editing, Supervision}

\cortext[cor1]{Corresponding Author}

\fntext[1]{}

\nonumnote{}

\nonumnote{}

\begin{abstract}
Safe motion planning in advanced driver-assistance systems and autonomous vehicles requires an accurate understanding of how the surrounding traffic scene is likely to evolve. However, many existing lane-change prediction methods remain centered on a single target vehicle, while multi-agent forecasting approaches often describe scene evolution only through future positions and provide limited explicit information about the maneuver associated with each vehicle. This study proposes a dynamic scene graph attention framework that predicts the lane-change intention and future trajectory of every relevant vehicle within a local traffic scene. The scene is represented as a time-varying interaction graph in which vehicles are modeled as nodes and their spatial and kinematic relationships are encoded through explicit edge features. Temporal graph-attention message passing captures evolving inter-vehicle dependencies and pre-maneuver cues, while an intention-guided decoder links each predicted maneuver to its corresponding future motion. A scene-level consistency objective further encourages compatible multi-vehicle futures. Experiments on the NGSIM I-80, NGSIM US-101, and highD datasets demonstrate consistent improvements over competing baselines. DSiGAT achieves intention prediction accuracies of 90.12\% and 90.97\% on NGSIM I-80 and US-101, respectively, and reduces trajectory RMSE by up to 52.94\% relative to the strongest baseline. It also produces lower inter-agent collision rates and joint displacement errors, indicating more coherent scene-level predictions. Ablation, sensitivity, robustness, and qualitative analyses further validate the contribution of the proposed components and the effectiveness of the scene-focused formulation.
\end{abstract}


% \begin{highlights}
% \item Proposes DSiGAT, a scene-level dynamic graph attention framework
% that jointly predicts the lane-change intentions and future trajectories
% of multiple interacting vehicles under explicit scene-level reasoning.
% \item Achieves superior performance on NGSIM I-80, NGSIM US-101, and highD, reaching up to 90.97\% intention accuracy and reducing trajectory RMSE by up to 52.94\% relative to the strongest baseline.
% \item Improves scene-level compatibility and deployment efficiency, delivering lower inter-agent collision rate and lightweight computation with only 1.02M parameters, 20.2 MB GPU memory, and 1.29 GFLOPs.
% \end{highlights}

%\nocite{*}

% Keywords
\begin{keywords}
lane-change intention prediction \sep trajectory forecasting \sep dynamic scene graph \sep graph attention \sep autonomous driving
\end{keywords}

\maketitle

\section{Introduction}

Advanced driver-assistance systems (ADAS) and autonomous vehicles (AVs) are expected to anticipate how the surrounding traffic environment is likely to evolve in order to make safe and timely driving decisions \cite{sharma2022introduction}. Although perception systems describe the current traffic scene \cite{asamoah2026querymotion}, planning requires an understanding of how that scene may change over the next few seconds \cite{hagedorn2024integration,kim2021stfp}. This is challenging in mixed traffic, where several nearby vehicles may act simultaneously and influence the actions available to the ego vehicle. A prediction focused on only one selected vehicle can provide only a partial view of the future traffic situation. Accordingly, reliable planning requires the behavior of the reference vehicle and all relevant surrounding vehicles to be considered within the same local traffic scene \cite{joseph2021autonomous,schulzinteraction,benterki2020artificial}.

Within this evolving traffic scene, lane changing is one of the most important behaviors for ADAS and AVs to anticipate \cite{do2023lane}. A lane change affects more than the vehicle performing the maneuver. It changes lane occupancy, available gaps, and leader--follower relationships, which may alter the behavior of several nearby vehicles \cite{wang2021intelligent}. The resulting traffic response is distributed across the local scene rather than confined to a single vehicle. Lane-change prediction must therefore account for the future behavior of the reference vehicle and the surrounding vehicles that may influence or respond to the maneuver \cite{lu2025lane,song2021surrounding}.

Existing lane-change prediction methods have achieved substantial progress by using vehicle kinematics \cite{das2021machine}, surrounding traffic information \cite{su2018learning}, and learned interaction features \cite{devi2026explainable,lu2025lane} to anticipate future maneuvers. Early approaches relied on probabilistic models and handcrafted descriptors, while more recent studies have adopted machine learning and deep learning architectures, including recurrent \cite{izquierdo2017vehicle,li2021lane,xue2022integrated}, convolutional \cite{shen2022parkpredict+,xie2020motion}, graph-based \cite{li2019grip,li2019grip++,wu2022lane,sheng2022graph}, and transformer models \cite{izquierdo2017vehicle,hao2020attention,xie2020motion,wang2023lane,liu2024laformer}. Despite these advances, many methods are still formulated around a predefined target vehicle. The surrounding vehicles are used to provide contextual information, but their own future behaviors are not predicted within the same scene. This target-centered formulation limits the model’s ability to represent how multiple vehicles may evolve together within the local traffic scene. To move beyond target-centered prediction, recent trajectory forecasting studies have increasingly adopted multi-agent formulations. These approaches model several vehicles within the same traffic scene and learn how their future motions are shaped by surrounding interactions. Social pooling \cite{deo2018convolutional}, graph neural networks \cite{wang2024socialformer}, and transformer-based scene encoders \cite{li2019grip++,wu2022lane} have been widely used to represent these dependencies and improve forecasting in dense traffic environments. This shift has enabled a broader view of scene evolution by predicting the motions of multiple interacting vehicles rather than treating each vehicle in isolation. However, the resulting scene representation is still commonly centered on future positions, with limited explicit characterization of the lane-change behavior of each vehicle.

Although multi-agent forecasting has expanded prediction from individual vehicles to the broader traffic scene, most existing scene representations remain centered on future positions. This limits their ability to explain how the scene is expected to evolve during lane-changing interactions. A predicted trajectory indicates where a vehicle may move, but it does not explicitly identify whether the vehicle intends to keep its lane, change left, or change right. At the scene level, this missing behavioral information is critical, since the maneuver of one vehicle can alter the available gaps, motion constraints, and likely responses of several neighboring vehicles. The central challenge is therefore to represent the future traffic scene in a form that explicitly captures both the lane-change behavior and corresponding motion of every relevant vehicle within the local traffic scene.

To address these limitations, this paper proposes a \emph{Dynamic Scene Interaction Graph Attention Network} (DSiGAT) for scene-level joint lane-change intention and trajectory prediction in mixed traffic. Instead of selecting one designated target vehicle and treating all others as passive context, the proposed framework represents the local traffic scene as a dynamic interaction graph in which vehicles are modeled as nodes and their time-varying relationships are encoded as directed edges. Graph attention-based message passing is then used to propagate interaction information across the scene and to learn representations that reflect each vehicle's behavior in the context of all others. Based on these shared scene representations, the framework predicts both lane-change intention and future trajectory for all valid vehicles in the scene. The proposed framework also introduces an explicit coupling between maneuver prediction and motion forecasting. For each vehicle, the predicted intention distribution is used to guide the trajectory decoder so that the predicted maneuver and the predicted future path remain semantically aligned. In addition, the framework incorporates a scene-level physical consistency constraint during training to discourage mutually incompatible futures across vehicles. This design moves beyond vehicle-wise prediction accuracy alone and instead aims to produce scene-level forecasts that are both informative and coherent.

Thus, the main contributions of this work are summarized as follows:
\begin{enumerate}
\item Lane-change prediction is formulated as a scene-level multi-agent problem in which the maneuver intention and future trajectory of all valid vehicles in a local traffic scene are predicted within a unified framework.

\item A dynamic scene interaction graph is developed to represent time-varying spatial and motion relationships among vehicles through directed graph construction and graph attention-based message passing.

\item An intention-guided trajectory decoding mechanism is introduced to associate the predicted maneuver of each vehicle with its corresponding future motion.

\item A scene-level consistency constraint is incorporated to penalize mutually incompatible future trajectories and promote physically coherent multi-vehicle predictions.

\end{enumerate}

The remainder of this paper is organized as follows. Section \ref{review} reviews the related literature. Section \ref{sec:data_processing} describes the data-processing and scene-construction pipeline. Section \ref{method} presents the proposed DSiGAT framework and training objective. Section \ref{experiment} introduces the datasets, experimental protocol, and evaluation metrics. Section \ref{results} presents the quantitative, qualitative, ablation, robustness, sensitivity, scene-level, and efficiency analyses. Section \ref{conclusion} concludes the paper.

\section{Literature Review}
\label{review}
In this section, we review the main lines of research relevant to the proposed work. We first examine lane-change intention prediction methods, which aim to classify future maneuver decisions of surrounding vehicles. We then discuss vehicle trajectory prediction approaches, with particular attention to multi-agent models that account for inter-vehicle interaction. Finally, we review joint intention--trajectory prediction frameworks and use this discussion to identify the remaining gap in scene-level prediction for multiple interacting vehicles.

\subsection{Lane-Change Intention Prediction}

Lane-change intention prediction was initially formulated as a decision or classification problem for a selected vehicle. Early studies relied on interpretable behavioral rules and handcrafted driving features. For example, \cite{gipps1986model} described lane changing through a rule-based behavioral model, while \cite{mandalia2005using} and \cite{kumar2013learning} used support vector machines to classify future maneuvers. Kumar et al. further combined multiclass support vector machine outputs with Bayesian filtering to support online prediction in highway traffic. Driver-monitoring studies also examined cues that appear before observable vehicle motion. \cite{doshi2009roles} showed that gaze and head dynamics can reveal maneuver intention earlier than vehicle kinematics alone, while \cite{kim2017prediction} demonstrated the value of augmented onboard sensing for intention recognition. Similarly, \cite{liu2020driving} incorporated surrounding-vehicle mobility into a hidden Markov model. Although these methods provided interpretable predictions, their dependence on manually selected features limited their ability to represent long temporal patterns and complex interactions.

The availability of larger naturalistic driving datasets shifted the field toward data-driven temporal modeling. \cite{lee2017convolution} applied a convolutional neural network to simplified bird's-eye-view traffic representations, while \cite{xing2020ensemble} used an ensemble BiLSTM to integrate driver, vehicle, and traffic information. \cite{shi2021improved} addressed class imbalance in LSTM-based intention prediction, and \cite{wang2021bayesian} introduced an adaptive LSTM with Bayesian threshold adjustment for changing traffic conditions. These studies improved the representation of temporal dependencies and reduced reliance on manually defined decision rules.

More recent approaches have incorporated attention, transformers, and graph-based representations to capture richer contextual information. \cite{ren2025lane} combined XGBoost and LSTM components with traffic-level and vehicle-type features. \cite{gao2023dual} proposed a dual-Transformer architecture for coupled intention and trajectory prediction, while \cite{liu2024lane} used a Transformer to model eye-tracking, preceding-vehicle, and oncoming-traffic cues. \cite{huang2024driver} introduced a topological graph with DGCN-SAM to represent driver behavior and traffic context, and \cite{zhou2025research} developed a Transformer-BiGRU framework for enhanced temporal modeling.

These developments have substantially improved the recognition of lane-change behavior by incorporating longer observation histories and broader traffic context. Nevertheless, the prediction output is commonly associated with one predefined target vehicle. Surrounding vehicles contribute contextual evidence for that prediction, but their own future maneuver states are generally not estimated within the same scene. Lane-change intention prediction has thus become increasingly context-aware while remaining predominantly target-centered.

\subsection{Vehicle Trajectory Prediction and Multi-Agent Motion Forecasting}

Trajectory prediction extends behavioral anticipation from identifying a maneuver to estimating how vehicle motion will evolve over time. Early approaches commonly predicted one agent at a time using kinematic assumptions or recurrent sequence models. This formulation was gradually expanded to include interaction information after it became clear that future motion depends on the surrounding traffic. An important step in this direction was Social LSTM, in which \cite{alahi2016social} introduced social pooling to capture dependencies among neighboring agents. \cite{gupta2018social} extended interaction-aware forecasting through Social GAN, which generated multimodal and socially acceptable trajectories using adversarial learning. These studies established that motion forecasting benefits from representing the behavior of nearby agents rather than processing each trajectory independently.

As trajectory forecasting became more closely connected to autonomous driving, models began to incorporate heterogeneous agents, vehicle dynamics, maps, and multiple possible futures. \cite{salzmann2020trajectron++} proposed Trajectron++, a graph-structured recurrent framework that combined agent interactions with dynamics and map information. \cite{phan2020covernet} formulated multimodal prediction as classification over a predefined set of feasible trajectories. \cite{chai2019multipath} and \cite{varadarajan2022multipath++} developed anchor-based multimodal forecasting through MultiPath and MultiPath++, demonstrating the value of representing uncertainty through multiple candidate futures. A related line of research emphasized structured scene encoding. \cite{gao2020vectornet} introduced VectorNet to represent agent histories and high-definition maps as vectorized elements. \cite{liang2020learning} proposed LaneGCN, which modeled lane topology and actor--map interactions through lane graphs. These methods expanded trajectory prediction beyond vehicle histories by integrating roadway structure and relational scene information.

More recent studies have increasingly adopted transformer-based scene models to capture long-range temporal and inter-agent dependencies. \cite{yuan2021agentformer} proposed AgentFormer, which jointly modeled temporal and social relationships through agent-aware attention. \cite{ngiam2021scene} introduced Scene Transformer as a unified architecture for predicting multiple agent trajectories within the same scene and addressed inconsistencies that may arise when agents are forecast independently. This scene-level direction was further advanced by HiVT \cite{zhou2022hivt}, Wayformer \cite{nayakanti2022wayformer}, MTR \cite{shi2022motion}, and MTR++ \cite{shi2024mtr++}. These models combined hierarchical attention, vectorized scene representations, and motion-query decoding to improve multimodal forecasting across multiple agents.

Multi-agent forecasting has consequently established a strong foundation for representing the future motion of several interacting vehicles within a common scene. The predicted scene, however, is usually described through positions, trajectories, latent motion modes, anchors, or query embeddings. Such representations indicate where vehicles may move, but they do not always provide an explicit lane-change state for each predicted vehicle. The behavioral meaning of the forecast is therefore often inferred from the resulting motion rather than produced directly as part of the scene prediction.

\subsection{Joint Intention--Trajectory Prediction}

Joint intention--trajectory prediction connects maneuver semantics with future motion. Instead of treating intention classification and trajectory forecasting as unrelated outputs, these methods use maneuver information to structure or guide future motion prediction. Early joint approaches were largely maneuver-conditioned. \cite{schreier2016integrated} used a Bayesian network to infer high-level maneuver probabilities and propagated maneuver-specific motion models to generate probabilistic trajectories. \cite{deo2018would} presented a freeway prediction framework that combined maneuver classification with class-conditional trajectory estimation. Deep sequence models later replaced manually specified maneuver modules. \cite{deo2018convolutional} proposed an LSTM encoder--decoder with convolutional social pooling and maneuver-conditioned multimodal trajectory outputs. \cite{xin2018intention} used two LSTM components, with the first recognizing lane-change intention and the second predicting long-horizon trajectories.

Later studies developed more explicit multi-task formulations. \cite{sui2021joint} proposed a multi-input Transformer that jointly predicted intention and trajectory for vulnerable road users using motion and scene information. For vehicle prediction, \cite{yuan2023temporal} introduced TMMOE, in which a shared temporal convolutional layer supplied task-specific experts for longitudinal motion, lateral motion, and driving intention under an uncertainty-weighted objective. \cite{deo2022multimodal} developed an MMAE framework that inferred lane-change intention from multiple spline-based hypotheses and updated future paths online through recursive least-squares estimation.

Recent methods have introduced richer mechanisms for relating maneuver hypotheses to future motion. \cite{do2023lane} conditioned multimodal forecasting on lane-graph traversals and used route hypotheses as a structured indication of intention. \cite{peng2025unipredictor} proposed UniPredictor, which combined a spatial--temporal anchor-attention encoder, a proposal-fusion intention decoder, and an autoregressive trajectory decoder. \cite{lu2026knowlcp} introduced KnowLCP, a knowledge-augmented dual-Transformer with parallel intention and trajectory branches supported by risk, kinematic, and intention-guidance cues. \cite{chen2025multi} moved toward a more scene-aware formulation through structured latent variables and an auxiliary lane-sequence task that aligned vehicle behavior with scene constraints, although the maneuver was represented through lane-level motion semantics rather than an explicit lane-change class. These methods demonstrate that maneuver information can improve the interpretation and prediction of future motion. Their principal output, however, is commonly generated for one target vehicle or for individually processed agents, even when surrounding traffic is encoded. Joint prediction in this context primarily refers to coupling two outputs for a vehicle rather than providing an explicit behavioral forecast for every relevant vehicle in the surrounding scene.

A related but distinct research direction addresses consistency among multiple predicted trajectories. \cite{luo2023jfp} proposed JFP, which performed joint future inference over multiple agents through a dynamic interaction graph and pairwise compatibility terms. The method reduced implausible overlaps by evaluating the compatibility of candidate futures across agents. This work demonstrated the importance of considering the predicted scene as a connected outcome rather than as a collection of independent trajectories. Its outputs, however, remained trajectory-based and did not explicitly assign lane-change states to the predicted vehicles.

\subsection{Research Gap and Motivation for Behavior-Aware Scene Prediction}
Existing studies have made substantial progress in lane-change intention prediction, multi-agent trajectory forecasting, and joint intention--trajectory modeling. However, these research directions remain only partially connected at the scene level. Intention prediction methods commonly provide explicit maneuver labels for a selected target vehicle, while multi-agent forecasting methods predict the future motion of several vehicles but often represent lane-change behavior only implicitly. Joint prediction models connect maneuver intention with trajectory forecasting, yet they are still frequently centered on individual vehicles rather than the complete local traffic scene. The remaining challenge is to explicitly predict the lane-change intention and future trajectory of every relevant vehicle within a local traffic scene while preserving consistency among their predicted motions. This is important because individually plausible predictions may become incompatible when combined, leading to conflicting lane occupancy, unsafe spacing, or unrealistic responses among interacting vehicles. These limitations motivate a behavior-aware scene forecasting framework that models the evolving interactions among vehicles and produces explicit, coherent predictions for the entire local traffic scene.

\section{Data Processing and Scene Construction}
\label{sec:data_processing}

Trajectory data from the Next Generation Simulation (NGSIM) and highD datasets are used to construct the local traffic scenes considered in this study. Both datasets provide detailed vehicle trajectories from multi-lane highway environments, although their variables and measurement formats differ. DSiGAT treats the local traffic scene, rather than an isolated target vehicle, as the basic prediction unit. Each scene contains the observed motion histories of multiple interacting vehicles and provides an intention label and a future trajectory for every valid vehicle. The raw trajectories are processed through four stages: trajectory cleaning, temporal-window extraction, local scene construction, and scene-level tensor assembly. The vehicle-level variables available in the NGSIM dataset, together with their descriptions and measurement units, are summarized in Table~\ref{tab:ngsim_vars}.

\begin{table}[t]
\centering
\caption{Vehicle-level variables in the NGSIM dataset}
\label{tab:ngsim_vars}
\begin{tabular}{l l c}
\toprule
\textbf{Variable} & \textbf{Description} & \textbf{Unit} \\
\midrule
Lane ID & Lane index & -- \\
Frame ID & Observation frame index & 0.1 s \\
Vehicle ID & Vehicle identifier & -- \\
Local X & Longitudinal position & ft \\
Local Y & Lateral position & ft \\
Vehicle Velocity & Vehicle speed & ft/s \\
Vehicle Acceleration & Vehicle acceleration & ft/s$^{2}$ \\
Preceding Vehicle & Preceding vehicle ID & -- \\
Following Vehicle & Following vehicle ID & -- \\
\bottomrule
\end{tabular}
\end{table}

\subsection{Trajectory Cleaning}
\label{subsec:trajectory_cleaning}

The raw NGSIM and highD trajectories are first screened to remove records
that cannot provide reliable observation and prediction sequences.
Table~\ref{tab:scene_filtering} summarizes the filtering rules.

For each vehicle, the visible duration is computed from its first and last
recorded frames. Since all trajectories are processed at $10\,\mathrm{Hz}$,
the minimum duration requirement of $7\,\mathrm{s}$ corresponds to
70 consecutive frames.

The original NGSIM positions, velocities, and accelerations are converted
from feet-based units to metres, metres per second, and metres per second
squared before any feature is computed. highD measurements are retained
in their original SI units. All continuous input features are subsequently
standardized using only the training-set statistics:
\begin{equation}
\widetilde{x}
=
\frac{x-\mu_{\mathrm{train}}}
{\sigma_{\mathrm{train}}},
\label{eq:feature_standardization}
\end{equation}
where $\mu_{\mathrm{train}}$ and $\sigma_{\mathrm{train}}$ denote the
training-set mean and standard deviation of the corresponding feature.
The same transformation is applied to the validation and test sets.

\begin{table*}[!t]
\centering
\caption{Trajectory filtering rules applied before scene construction.}
\label{tab:scene_filtering}
\renewcommand{\arraystretch}{1.10}
\setlength{\tabcolsep}{5pt}
\begin{tabular}{p{3.2cm} p{5.3cm} p{6.0cm}}
\toprule
\textbf{Category}
& \textbf{Condition}
& \textbf{Reason for exclusion} \\
\midrule

Short-duration track
& The total visible duration is less than $7\,\mathrm{s}$.
& The track cannot provide the complete $3\,\mathrm{s}$ observation
history and $4\,\mathrm{s}$ future trajectory. \\

Boundary-truncated track
& For NGSIM, the first or last recorded position lies within approximately
$50\,\mathrm{m}$ of the upstream or downstream boundary of the monitored
freeway segment.
& The vehicle is entering or leaving the fixed camera coverage, and its
surrounding traffic context may therefore be incomplete. \\

Unstable lane record
& The lane-index sequence contains two lane changes in opposite directions
within less than $5\,\mathrm{s}$.
& Rapid lane-index reversals are likely to represent annotation noise,
irregular ramp movements, or behavior outside the lane-change setting
considered in this study. \\

Non-mainline NGSIM record
& The NGSIM lane index satisfies $\ell_i^t \geq 7$.
& These indices correspond to auxiliary, ramp, or non-mainline roadway
regions and are not included in the freeway lane-change analysis. \\

Observation-window lane change
& A persistent lane-index transition begins during the $3\,\mathrm{s}$
observation interval.
& The maneuver would already be partially visible in the model input,
which would weaken the intended anticipation setting. \\

\bottomrule
\end{tabular}
\end{table*}

\subsection{Temporal-Window Extraction}
\label{subsec:window_extraction}

After cleaning, the trajectories are converted into fixed-length temporal
samples. Each sample spans $7\,\mathrm{s}$ and contains a $3\,\mathrm{s}$
observation interval followed by a $4\,\mathrm{s}$ prediction interval:
$T_{\mathrm{win}}
=
T_{\mathrm{obs}}+T_{\mathrm{pred}}
=
3\,\mathrm{s}+4\,\mathrm{s}
=
7\,\mathrm{s}.
$
At $10\,\mathrm{Hz}$, each sample therefore contains
$
F_{\mathrm{obs}}=30,
\qquad
F_{\mathrm{pred}}=40
$
observation and prediction frames, respectively.

A lane-change event time, denoted by $t_{\mathrm{LC}}$, is identified as
the first frame of a persistent transition from the current lane to an
adjacent lane after unstable lane-index reversals have been removed.
For anticipation time $T\in\{3,2,1,0\}\,\mathrm{s}$, the 3-second
observation window spans
$[t_{\mathrm{LC}}-T-3\,\mathrm{s},\,t_{\mathrm{LC}}-T]$.
At $T=0$, the onset frame itself is excluded from the input. This ensures that the lane change is a
future event rather than an action already observed by the model.

Lane-change scenes are extracted with a stride of $0.5\,\mathrm{s}$.
The denser stride captures the gradual development of pre-maneuver motion
and interaction cues. Lane-keeping scenes are extracted only from stable
trajectory segments that remain at least $5\,\mathrm{s}$ away from any
lane-change event. These scenes use a larger stride of $3\,\mathrm{s}$ to
avoid producing many nearly identical lane-keeping samples.

\subsection{Local Scene Construction}
\label{subsec:scene_construction}

Each temporal window is converted into a local multi-vehicle traffic scene.
A reference vehicle is first selected to define the center of the local
neighborhood. In a lane-change scene, the reference vehicle is one whose
lane change occurs during the prediction interval. In a lane-keeping scene,
the reference vehicle is selected from a stable trajectory segment located
at least $5\,\mathrm{s}$ from any lane-change event.

All vehicles that are continuously available over the complete
$3\,\mathrm{s}$ observation and $4\,\mathrm{s}$ prediction intervals are
considered candidate scene members. Candidates are ranked according to
their Euclidean distance from the reference vehicle in the
longitudinal--lateral plane at the final observation frame. The reference
vehicle and its nearest surrounding vehicles are then retained.

Each scene contains at most
$
N_{\max}=6
$
vehicles, including the reference vehicle. This limit preserves the most
relevant local interactions while keeping graph message passing and
pairwise scene-level evaluation computationally manageable. If more than
6 valid vehicles are available, only the 6 nearest vehicles are kept.
If fewer than 6 vehicles are available, the remaining entries are padded
with zeros. A binary validity mask
$
\mathbf{M}_s \in \{0,1\}^{N_{\max}}
$
identifies real vehicles and padded entries so that padded nodes do not
participate in graph construction, attention, loss computation, or
evaluation.

\subsection{Lane-Direction Normalization and Maneuver Labels}
\label{subsec:scene_labels}

Raw lane identifiers do not necessarily follow the same left-to-right
ordering across roadway directions and datasets. Before assigning maneuver
labels, the lane identifiers are therefore converted to a common
driving-direction-relative convention. Under this convention, a decrease
in the normalized lane index represents movement to the left, while an
increase represents movement to the right.

Let $\bar{\ell}_i^t$ denote the normalized lane index of vehicle $v_i$ at
frame $t$. Each vehicle is assigned one of three maneuver labels:
lane keeping (LK), left lane change (LLC), or right lane change (RLC).
The label is determined from the lane-index evolution during the
40-frame prediction interval. Let $\bar{\ell}_i^{30}$ be the lane index
at the final observation frame and let $\bar{\ell}_i^{30+m}$ be the lane
index at future step $m$. The maneuver label is
\begin{equation}
c_i =
\begin{cases}
\mathrm{LLC},
&
\text{if }
\bar{\ell}_i^{30+m}
<
\bar{\ell}_i^{30}
\text{ for a persistent future transition},\\[4pt]

\mathrm{RLC},
&
\text{if }
\bar{\ell}_i^{30+m}
>
\bar{\ell}_i^{30}
\text{ for a persistent future transition},\\[4pt]

\mathrm{LK},
&
\text{otherwise}.
\end{cases}
\label{eq:scene_maneuver_label}
\end{equation}

A transition is treated as persistent only when the vehicle remains in the
new lane for the remainder of the available prediction sequence. This
prevents short lane-index fluctuations from being labeled as genuine lane
changes. Because scenes containing lane changes during the observation
interval have already been removed, every LLC or RLC label corresponds to
a maneuver that begins after the observation period.

\subsection{Scene-Level Tensor Assembly}
\label{subsec:tensor_assembly}

Let $N_s \leq N_{\max}$ denote the number of valid vehicles in scene $s$.
For each vehicle $v_i$, the observed feature history is written as
$
\mathbf{X}_i
=
\left[
\mathbf{x}_i^1,
\mathbf{x}_i^2,
\ldots,
\mathbf{x}_i^{F_{\mathrm{obs}}}
\right],
$
$
\mathbf{x}_i^t \in \mathbb{R}^{d},
$
where $d$ is the per-frame node-feature dimension. Stacking the vehicle
histories produces the scene observation tensor
$
\mathbf{X}_s
\in
\mathbb{R}^{N_{\max}\times 30\times d}.
\label{eq:scene_input_tensor}
$

The future trajectory of each vehicle is expressed in a local coordinate
system anchored at that vehicle's position at the final observation frame.
For vehicle $v_i$, the target trajectory is
\begin{equation}
\mathbf{Y}_i
=
\left[
\mathbf{y}_i^1,
\mathbf{y}_i^2,
\ldots,
\mathbf{y}_i^{40}
\right],
\qquad
\mathbf{y}_i^m \in \mathbb{R}^{2},
\end{equation}
where each vector contains the longitudinal and lateral displacement at
future step $m$. Stacking all vehicle trajectories gives
$
\mathbf{Y}_s
\in
\mathbb{R}^{N_{\max}\times 40\times 2}.
\label{eq:scene_target_tensor}
$

The scene maneuver labels are collected as
\begin{equation}
\mathbf{c}_s
=
[c_1,c_2,\ldots,c_{N_{\max}}],
\end{equation}
with padded entries ignored through $\mathbf{M}_s$. The complete
scene-level sample is therefore
\begin{equation}
\mathcal{S}_s
=
\left(
\mathbf{X}_s,
\mathbf{Y}_s,
\mathbf{c}_s,
\mathbf{M}_s
\right).
\label{eq:complete_scene_sample}
\end{equation}

Accordingly, one model input contains the shared motion history of all
valid vehicles in a local traffic scene, while the corresponding targets
contain their future trajectories and maneuver intentions. This
representation allows DSiGAT to predict all valid vehicles jointly and to
evaluate whether their predicted futures remain mutually compatible.

\section{Methodology}
\label{method}

\subsection{Problem Formulation}

Lane-change prediction in real traffic is inherently a multi-agent problem. A vehicle may decide to keep its lane, change left, or change right depending on the surrounding gaps, the motion of leading and following vehicles, and the evolving traffic conditions in adjacent lanes. For this reason, the objective of this study is not to predict one selected vehicle in isolation, but to jointly predict the future behavior of all relevant vehicles within a local traffic scene.

Consider a traffic scene observed over an input horizon of $F_{\mathrm{obs}}$ frames. Let the scene contain $N$ valid vehicles, $V = \{v_1, v_2, \dots, v_N\}$. For each vehicle $v_i$, its observed history is represented by a sequence of feature vectors over the observation window, $\mathbf{X}_i = \left[\mathbf{x}_i^1, \mathbf{x}_i^2, \dots, \mathbf{x}_i^{F_{\mathrm{obs}}}\right]$, where $\mathbf{x}_i^t \in \mathbb{R}^d$ denotes the state of vehicle $v_i$ at frame $t$, and $d$ is the feature dimension. These features describe the recent motion and lane-related state of the vehicle. The full scene input is therefore written as $\mathbf{X} = \{\mathbf{X}_1, \mathbf{X}_2, \dots, \mathbf{X}_N\}$.

Given this observed scene history, the goal is to predict, for every vehicle $v_i$, both its future lane-change intention and its future trajectory over a prediction horizon of $F_{\mathrm{pred}}$ frames. The intention output is defined as a probability vector $\boldsymbol{\pi}_i = 
\left[\pi_i^{\mathrm{LK}}, \pi_i^{\mathrm{LLC}}, \pi_i^{\mathrm{RLC}}\right]^\top \in [0,1]^3$,  where $\pi_i^{\mathrm{LK}}$, $\pi_i^{\mathrm{LLC}}$, and $\pi_i^{\mathrm{RLC}}$ denote the probabilities that vehicle $v_i$ will keep its lane, change to the left lane, or change to the right lane, respectively. The trajectory output is defined as
$\hat{\mathbf{Y}}_i \in \mathbb{R}^{F_{\mathrm{pred}}\times 2}$,
where each row gives the predicted longitudinal and lateral
displacement of vehicle $v_i$ relative to its position at the
final observation frame.

The overall prediction task is therefore to learn a model $f_{\theta}$ such that
\begin{equation}
  f_{\theta}(\mathbf{X})
=
\left\{
(\boldsymbol{\pi}_1,\hat{\mathbf{Y}}_1),
(\boldsymbol{\pi}_2,\hat{\mathbf{Y}}_2),
\dots,
(\boldsymbol{\pi}_N,\hat{\mathbf{Y}}_N)
\right\},  
\end{equation}

where $\theta$ denotes the trainable parameters of the model. In other words, a single forward pass over the observed traffic scene produces intention and trajectory predictions for all valid vehicles in that scene.

This formulation differs from target-centered lane-change prediction, where one designated vehicle is treated as the only prediction subject and all other vehicles serve only as context. In the present work, every valid vehicle in the scene is treated as a prediction target. This allows the model to represent traffic behavior at the scene level and to reason directly about the coupled futures of interacting vehicles.

% In addition to per-vehicle accuracy, the predicted scene must remain physically and behaviorally coherent. First, the predicted intention of a vehicle should agree with its predicted trajectory. For example, a vehicle with a high left-lane-change probability should not produce a future path that remains fully centered in its current lane. Second, the predictions across different vehicles should be mutually compatible. Two vehicles should not be predicted to occupy the same space at the same future time step, and their joint futures should remain consistent with the traffic configuration from which they emerged. These requirements motivate a scene-level formulation in which intention prediction, trajectory prediction, and cross-vehicle consistency are learned together within one unified framework.

\subsection{Dynamic Scene Graph Representation}

After scene construction, each sample is represented as a time-varying interaction graph. This representation is used to model the local traffic scene as a set of vehicles whose states evolve over time and whose behaviors are shaped by directed pairwise interactions. Representing the scene in graph form allows the model to move beyond fixed neighbor slots and to learn which surrounding vehicles are most relevant at each moment.

\subsubsection{Node Representation}

Each retained vehicle is represented as a graph node at every observation frame. For a scene with $N_s$ valid vehicles, the node set at frame $t$ is written as
\begin{equation}
\mathcal{V}^t = \{v_1^t, v_2^t, \dots, v_{N_s}^t\}.
\end{equation}

The node feature vector of vehicle $v_i$ at frame $t$ is denoted by $\mathbf{x}_i^t$. It contains the per-vehicle quantities used to describe the recent motion state and lane status of that vehicle. In the current formulation, the node features include longitudinal speed, longitudinal acceleration, lateral acceleration, cumulative longitudinal displacement from the first observation frame, lateral displacement from the first observation frame, and lane identity encoded as a one-hot vector. Accordingly, the node feature vector is written as
\begin{equation}
\mathbf{x}_i^t =
\left[
v_{x,i}^t,\;
a_{x,i}^t,\;
a_{y,i}^t,\;
\Delta x_i^t,\;
\Delta y_i^t,\;
\mathbf{l}_i^t
\right],
\end{equation}
where $v_{x,i}^t$ is the longitudinal speed, $a_{x,i}^t$ and $a_{y,i}^t$ are the longitudinal and lateral accelerations, $\Delta x_i^t$ and $\Delta y_i^t$ are the cumulative longitudinal and lateral displacements measured relative to the first observation frame, and $\mathbf{l}_i^t$ is the one-hot lane encoding. Continuous features are standardized using statistics computed from the training data, while the lane encoding preserves lane identity without imposing an artificial ordinal relation between lanes.

This node design keeps the per-vehicle state representation compact and interpretable. Motion and lane information are stored directly in the node features, while interaction and safety-related information are introduced through the graph edges rather than being merged into each node.

\subsubsection{Edge Construction and Edge Features}

Directed edges are used to represent potential influence between vehicles. At each frame $t$, an edge from vehicle $v_j$ to vehicle $v_i$ is created when the two vehicles satisfy both of the following conditions: they are in the same lane or in adjacent lanes, and their absolute longitudinal separation is less than 100\,m. Under these conditions, vehicle $v_j$ is considered a relevant neighboring agent for vehicle $v_i$. The directed edge set at frame $t$ is therefore written as
\begin{equation}
\mathcal{E}^t = \left\{(v_j^t \rightarrow v_i^t)\; \middle|\; r_{ij}^t \leq 100\text{ m and } \delta_{ij}^t \in \{-1,0,1\}\right\},
\end{equation}
where $r_{ij}^t$ denotes the absolute longitudinal distance between vehicles $i$ and $j$, and $\delta_{ij}^t$ denotes their lane relation, with values corresponding to left-adjacent, same-lane, or right-adjacent configurations.

Each directed edge carries a feature vector that describes the pairwise relationship between the two vehicles. For the edge $(v_j^t \rightarrow v_i^t)$, the feature vector is defined as
\begin{equation}
\mathbf{e}_{ij}^t =
\left[
\Delta x_{ij}^t,\;
\Delta y_{ij}^t,\;
\Delta v_{x,ij}^t,\;
\Delta v_{y,ij}^t,\;
\mathbf{r}_{ij}^t,\;
\mathrm{TTC}_{ij}^t
\right],
\end{equation}
where $\Delta x_{ij}^t$ is the signed longitudinal gap, $\Delta y_{ij}^t$ is the lateral offset, $\Delta v_{x,ij}^t$ and $\Delta v_{y,ij}^t$ are the relative longitudinal and lateral speeds, $\mathbf{r}_{ij}^t$ is the categorical lane-relation encoding, and $\mathrm{TTC}_{ij}^t$ is an approximate time-to-collision value.

The time-to-collision term is included to provide a compact safety-oriented description of the interaction state. It is computed from the longitudinal separation and relative longitudinal speed and is written as
\begin{equation}
\mathrm{TTC}_{ij}^t =
\begin{cases}
\dfrac{|\Delta x_{ij}^t|}{\max(|\Delta v_{x,ij}^t|,\epsilon)}, & \text{if } \Delta x_{ij}^t \Delta v_{x,ij}^t < 0, \\[8pt]
T_{\max}, & \text{otherwise},
\end{cases}
\end{equation}
where $\epsilon$ is a small constant used to avoid division by zero, and $T_{\max}$ is a saturation value assigned when the two vehicles are not closing longitudinally. This term does not replace the full geometric interaction state, but it provides an additional safety-related cue that is directly relevant to lane-change feasibility and gap acceptance.

This edge design serves two purposes. First, it captures interaction quantities that cannot be represented by a vehicle alone, such as relative spacing and closing speed. Second, it avoids the rigid structure of fixed-slot neighbor encoding by allowing the set of active neighbors to change with the traffic scene.

\subsubsection{Temporal Graph Sequence}

The observation window contains 30 frames, and a graph is constructed at each of these frames. The full scene is therefore represented as a temporal graph sequence
\begin{equation}
\mathcal{G} =
\left\{
\mathcal{G}^1,\mathcal{G}^2,\dots,\mathcal{G}^{30}
\right\},
\qquad
\mathcal{G}^t = (\mathcal{V}^t,\mathcal{E}^t).
\end{equation}

The node set remains aligned with the retained vehicles in the scene, but both the node values and the edge structure may change from one frame to the next. As vehicles accelerate, decelerate, change lanes, or alter their relative spacing, the corresponding node features and edge features are updated accordingly. The graph is therefore dynamic not only in feature values but also in connectivity.

This temporal graph sequence is the direct input to the proposed DSiGAT framework. It preserves three complementary aspects of the observed traffic scene: the motion history of each vehicle through the node features, the pairwise interaction state through the edge features, and the time-varying scene structure through the graph sequence itself. These three components together provide the representation needed for joint scene-level intention and trajectory prediction.

\subsection{Overview of the Proposed Framework}
\begin{figure*}
    \centering
    \includegraphics[width=1\linewidth]{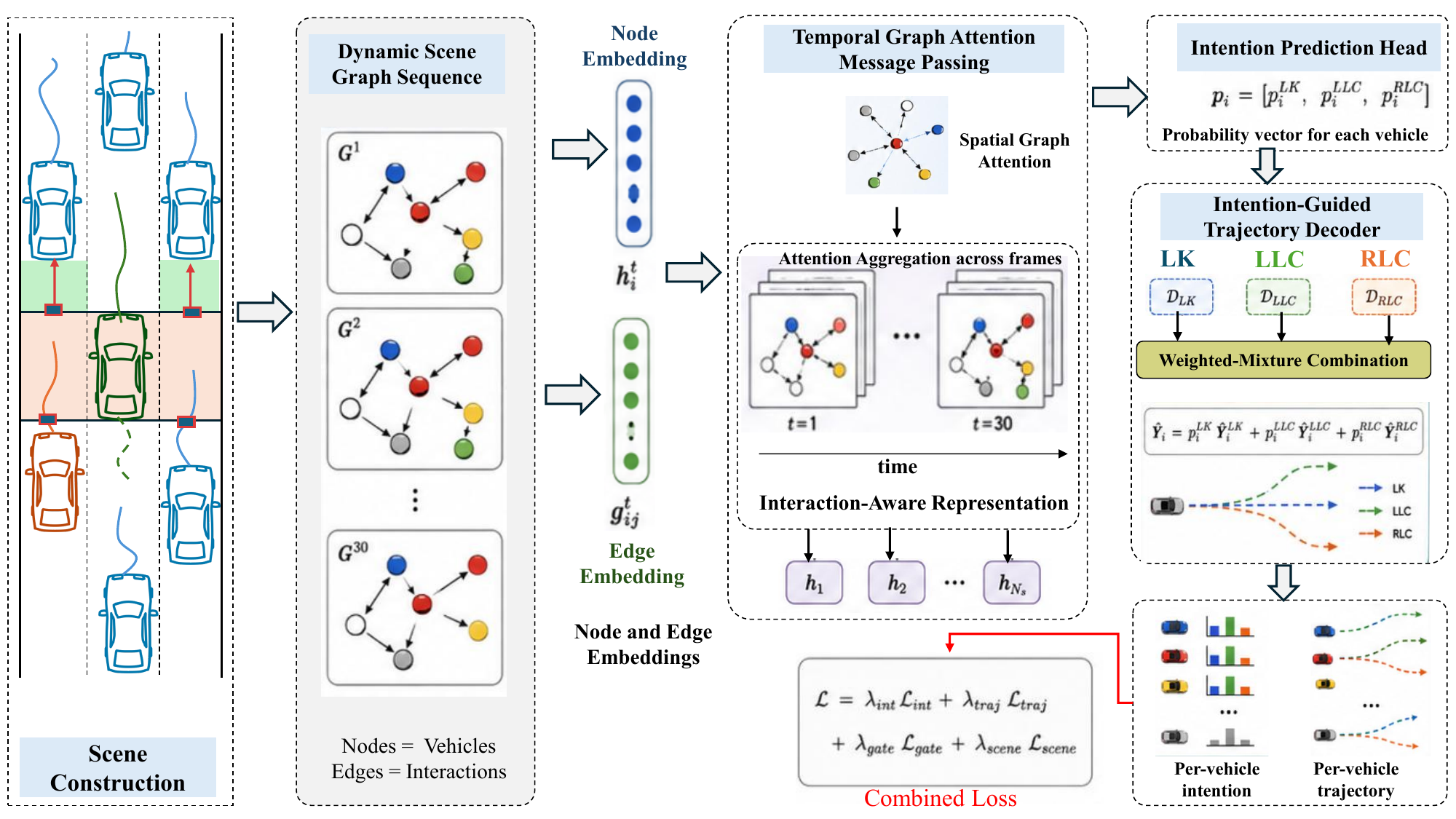}
    \caption{Overall architecture of DSiGAT for scene-level joint lane-change intention and trajectory prediction. Dynamic vehicle graphs are encoded through node-edge embeddings and temporal graph attention, followed by per-vehicle intention classification, intention-guided trajectory decoding, and joint loss optimization}
    \label{fig:overall}
\end{figure*}

Building on the dynamic scene graph representation, the proposed Dynamic Scene Interaction Graph Attention Network (DSiGAT) jointly predicts lane-change intention and future trajectory for all valid vehicles in a local traffic scene. As illustrated in Figure~\ref{fig:overall}, the model consists of four main components: node and edge embedding, temporal graph attention message passing, an intention prediction head, and an intention-guided trajectory decoder. In addition, a scene-level consistency constraint is introduced during training to discourage mutually incompatible futures across vehicles. The following subsections describe these components in detail.

\subsubsection{Node and Edge Embedding}

At each observation frame, the scene graph contains node features that describe the motion and lane state of each vehicle, and edge features that describe the pairwise interaction between connected vehicles. These raw features are not used directly for graph attention. Instead, they are first projected into learnable latent spaces so that node and edge information can be represented in a form suitable for message passing.

For a vehicle $v_i$ at frame $t$, let $\mathbf{x}_i^t \in \mathbb{R}^{d_n}$ denote its raw node feature vector. Similarly, for a directed edge $(v_j^t \rightarrow v_i^t)$, let $\mathbf{e}_{ij}^t \in \mathbb{R}^{d_e}$ denote its raw edge feature vector. The node features are embedded through a learnable projection function
\begin{equation}
\mathbf{h}_i^t = \phi_n(\mathbf{x}_i^t),
\qquad
\mathbf{h}_i^t \in \mathbb{R}^{d_h},
\end{equation}
where $\phi_n(\cdot)$ is implemented as a linear layer followed by layer normalization and a ReLU activation. In the present design, the node embedding dimension is set to $d_h = 128$.

The edge features are embedded in the same way through a separate projection function
\begin{equation}
\mathbf{g}_{ij}^t = \phi_e(\mathbf{e}_{ij}^t),
\qquad
\mathbf{g}_{ij}^t \in \mathbb{R}^{d_g},
\end{equation}
where $\phi_e(\cdot)$ is also implemented as a linear layer followed by layer normalization and ReLU. The edge embedding dimension is set to $d_g = 64$.

Using separate embedding functions for nodes and edges allows the model to preserve the difference between per-vehicle state information and pairwise interaction information. The node embeddings summarize how each vehicle is moving at a given frame, while the edge embeddings summarize how that vehicle is positioned and moving relative to its connected neighbors. These embedded features form the input to the graph attention module described next.

\subsubsection{Temporal Graph Attention Message Passing}
Lane-change behavior is shaped by both local vehicle interactions and their evolution over time. The purpose of this stage is to learn, for each vehicle, a representation that reflects not only its own motion history but also the influence of nearby vehicles across the observation window. To achieve this, DSiGAT performs graph attention-based message passing at each frame and then aggregates the resulting representations temporally.

At a given frame $t$, let $\mathbf{h}_i^t \in \mathbb{R}^{128}$ denote the embedded node feature of vehicle $v_i$, and let $\mathbf{g}_{ij}^t \in \mathbb{R}^{64}$ denote the embedded edge feature associated with the directed edge $(v_j^t \rightarrow v_i^t)$. For each connected pair, an attention score is computed to determine how strongly vehicle $v_j$ should influence vehicle $v_i$ at frame $t$. This score is defined as
\begin{equation}
e_{ij}^t
=
\mathbf{a}^{\top}
\left[
\mathbf{W}_q \mathbf{h}_i^t \; \Vert \;
\mathbf{W}_k \mathbf{h}_j^t \; \Vert \;
\mathbf{W}_e \mathbf{g}_{ij}^t
\right],
\end{equation}
where $\mathbf{W}_q$, $\mathbf{W}_k$, and $\mathbf{W}_e$ are learnable projection matrices, $\mathbf{a}$ is a learnable attention vector, and $\Vert$ denotes vector concatenation. The normalized attention coefficient is then obtained by applying a softmax over all incoming neighbors of vehicle $v_i$:
\begin{equation}
\alpha_{ij}^t
=
\frac{\exp(\mathrm{LeakyReLU}(e_{ij}^t))}
{\sum\limits_{k \in \mathcal{N}_i^t}
\exp(\mathrm{LeakyReLU}(e_{ik}^t))},
\end{equation}
where $\mathcal{N}_i^t$ denotes the set of vehicles connected to $v_i$ at frame $t$.

Using these coefficients, the node representation of vehicle $v_i$ is updated as
\begin{equation}
\tilde{\mathbf{h}}_i^t
=
\mathrm{ReLU}
\left(
\mathbf{W}_s \mathbf{h}_i^t
+
\sum_{j \in \mathcal{N}_i^t}
\alpha_{ij}^t
\mathbf{W}_m
\left[
\mathbf{h}_j^t \; \Vert \; \mathbf{g}_{ij}^t
\right]
\right),
\end{equation}
where $\mathbf{W}_s$ and $\mathbf{W}_m$ are learnable weight matrices. This update combines self-information with interaction messages received from neighboring vehicles. In the proposed framework, this graph attention operation is performed using 8 attention heads and repeated for 3 message-passing rounds at each frame. This allows each vehicle representation to progressively absorb richer local interaction structure. 

The frame-level graph features obtained for vehicle $v_i$ over the 30-frame observation window are then written as
$\left\{
\tilde{\mathbf{h}}_i^1,
\tilde{\mathbf{h}}_i^2,
\dots,
\tilde{\mathbf{h}}_i^{30}
\right\}$.
These features are not equally informative across time, since some moments carry stronger pre-maneuver cues than others. To summarize the observation window more effectively, a temporal attention mechanism is used to assign higher weight to the most informative frames. The temporal attention score at frame $t$ is computed as
\begin{equation}
u_i^t
=
\mathbf{v}^{\top}
\tanh
\left(
\mathbf{W}_t \tilde{\mathbf{h}}_i^t + \mathbf{b}_t
\right),
\end{equation}
where $\mathbf{W}_t$, $\mathbf{b}_t$, and $\mathbf{v}$ are learnable parameters. The normalized temporal weight is then
\begin{equation}
\beta_i^t
=
\frac{\exp(u_i^t)}
{\sum\limits_{\tau=1}^{30} \exp(u_i^{\tau})}.
\end{equation}
The final graph-aware representation of vehicle $v_i$ is obtained by the weighted temporal aggregation
\begin{equation}
\mathbf{h}_i
=
\sum_{t=1}^{30}
\beta_i^t \tilde{\mathbf{h}}_i^t.
\end{equation}

\subsubsection{Intention Prediction Head}

The graph attention module produces one scene-aware representation for each retained vehicle. These representations are then mapped to maneuver probabilities through the intention prediction head. For vehicle $v_i$, let $\mathbf{h}_i \in \mathbb{R}^{d_h}$ denote its final graph-aware representation. The intention head applies a two-layer feedforward network to $\mathbf{h}_i$:
\begin{equation}
\mathbf{z}_i = \mathrm{ReLU}\!\left(\mathbf{W}_1 \mathbf{h}_i + \mathbf{b}_1\right),
\end{equation}
followed by layer normalization and dropout, and then computes the output logits as
\begin{equation}
\mathbf{o}_i = \mathbf{W}_2 \mathbf{z}_i + \mathbf{b}_2,
\end{equation}
where $\mathbf{o}_i \in \mathbb{R}^{3}$.

The maneuver probability vector is obtained by applying the softmax function:
\begin{equation}
\boldsymbol{\pi}_i
=
\mathrm{softmax}(\mathbf{o}_i)
=
\left[
\pi_i^{\mathrm{LK}},
\pi_i^{\mathrm{LLC}},
\pi_i^{\mathrm{RLC}}
\right]^\top,
\end{equation}
where $\pi_i^{\mathrm{LK}}$, $\pi_i^{\mathrm{LLC}}$, and $\pi_i^{\mathrm{RLC}}$ denote the predicted probabilities of lane keeping, left lane change, and right lane change, respectively.

This produces one maneuver distribution for each retained vehicle in the scene. The predicted intention is used both as an explicit output of the model and as a conditioning signal for the trajectory decoder described next.

\subsubsection{Intention-Guided Trajectory Decoder}

The predicted maneuver distribution should not remain separate from the trajectory prediction stage. A vehicle that is likely to keep its lane should produce a different future path from one that is likely to move left or right. For this reason, the trajectory decoder is conditioned directly on the predicted intention probabilities so that maneuver prediction and future motion remain aligned within the same vehicle.

Let $\mathbf{h}_i \in \mathbb{R}^{d_h}$ denote the graph-aware representation of vehicle $v_i$, and let
\[
\boldsymbol{\pi}_i
=
\left[
\pi_i^{\mathrm{LK}},
\pi_i^{\mathrm{LLC}},
\pi_i^{\mathrm{RLC}}
\right]^\top
\]
denote its predicted intention distribution. Instead of using a single decoder for all maneuver types, the proposed framework maintains three maneuver-specific trajectory decoders: one for lane keeping, one for left lane change, and one for right lane change. Each decoder maps $\mathbf{h}_i$ to a trajectory prototype over the prediction horizon:
\begin{equation}
\begin{aligned}
\hat{\mathbf{Y}}_i^{\mathrm{LK}} &= D_{\mathrm{LK}}(\mathbf{h}_i), \\
\hat{\mathbf{Y}}_i^{\mathrm{LLC}} &= D_{\mathrm{LLC}}(\mathbf{h}_i), \\
\hat{\mathbf{Y}}_i^{\mathrm{RLC}} &= D_{\mathrm{RLC}}(\mathbf{h}_i).
\end{aligned}
\end{equation}
where each decoder output lies in $\mathbb{R}^{40 \times 2}$.

Each decoder is implemented as a two-layer feedforward network whose final output is reshaped into a 40-step sequence of longitudinal and lateral displacements. The final predicted trajectory is then obtained as a weighted combination of the three maneuver-specific outputs:
\begin{equation}
\hat{\mathbf{Y}}_i
=
\pi_i^{\mathrm{LK}} \hat{\mathbf{Y}}_i^{\mathrm{LK}}
+
\pi_i^{\mathrm{LLC}} \hat{\mathbf{Y}}_i^{\mathrm{LLC}}
+
\pi_i^{\mathrm{RLC}} \hat{\mathbf{Y}}_i^{\mathrm{RLC}}.
\end{equation}

This formulation allows the trajectory prediction to reflect the model's own maneuver belief. When the intention prediction is confident, the final trajectory is dominated by the corresponding maneuver-specific decoder. When the maneuver prediction is uncertain, the final trajectory becomes a softer combination of alternative future motion patterns. In both cases, the trajectory remains explicitly linked to the predicted intention distribution. The decoder therefore produces one future trajectory $\hat{\mathbf{Y}}_i \in \mathbb{R}^{40 \times 2}$ for each retained vehicle in the scene. These trajectories are used both for trajectory supervision and for the scene-level consistency constraint introduced next. 

Although the trajectory decoder predicts vehicle-relative displacements, 
scene-level compatibility must be evaluated in a common spatial coordinate 
system. Let $\mathbf{q}_{i}^{30}$ denote the position of vehicle $v_i$ at 
the final observation frame in the shared scene coordinate system. The 
predicted absolute position at future step $m$ is reconstructed as
\begin{equation}
    \hat{\mathbf{p}}_{i}^{m}
    =
    \mathbf{q}_{i}^{30}
    +
    \hat{\mathbf{Y}}_{i}^{m}.
\end{equation}
The corresponding ground-truth position is obtained in the same manner. 
These reconstructed scene-coordinate positions are used in the scene-level 
consistency loss, IACR, and JDE calculations.

\subsubsection{Scene-Level Consistency Constraint}

The intention-guided decoder produces one future trajectory for each retained vehicle in the scene. These trajectories should be accurate at the vehicle level, but they should also remain mutually compatible when viewed together as a future traffic scene. To encourage this, a scene-level consistency constraint is introduced during training to penalize predicted futures in which two vehicles move into physically incompatible positions.

Consider two retained vehicles, $v_i$ and $v_j$, with predicted future positions at prediction step $m$ denoted by $\hat{\mathbf{p}}_i^m$ and $\hat{\mathbf{p}}_j^m$, respectively. A collision penalty is applied when the separation between the two predicted positions falls below a predefined safety threshold. The pairwise penalty at step $m$ is written as
\begin{equation}
\ell_{ij}^{m}
=
\max
\left(
0,\;
D_{\mathrm{safe}} - \left\lVert \hat{\mathbf{p}}_i^m - \hat{\mathbf{p}}_j^m \right\rVert_2
\right),
\end{equation}
where $D_{\mathrm{safe}}$ denotes the minimum allowable separation distance.

This penalty is evaluated for all distinct vehicle pairs and for all prediction steps. The resulting scene-level consistency loss is defined as
\begin{equation}
L_{\mathrm{scene}}
=
\frac{1}{N_s (N_s - 1) F_{\mathrm{pred}}}
\sum_{i \neq j}
\sum_{m=1}^{F_{\mathrm{pred}}}
\ell_{ij}^{m},
\end{equation}
where $N_s$ is the number of valid vehicles in the scene and $F_{\mathrm{pred}} = 40$ is the prediction horizon.

This loss becomes zero when all predicted trajectories remain sufficiently separated throughout the forecast interval. Its value increases when the model produces futures in which different vehicles approach each other too closely or overlap in space and time. In this way, the constraint encourages the model to generate scene-level predictions that remain physically plausible rather than only accurate on a vehicle-by-vehicle basis.

The scene-level term complements the per-vehicle intention and trajectory objectives. The intention loss guides maneuver classification, the trajectory loss guides future motion prediction, and the scene-level consistency term discourages incompatible joint futures across vehicles. The complete training objective that combines these components is described next.

\subsubsection{Training Objective}

The proposed framework is trained end-to-end by jointly optimizing maneuver classification, trajectory regression, and scene-level consistency. Let $N_s$ denote the number of valid vehicles in scene $s$. For each retained vehicle $v_i$, the model predicts an intention distribution $\boldsymbol{\pi}_i$ and a future trajectory $\hat{\mathbf{Y}}_i$. The total loss combines these objectives so that the model learns accurate per-vehicle predictions while also preserving compatibility across the full scene.

The intention prediction loss is defined as the average cross-entropy over all valid vehicles in the scene:
\begin{equation}
L_{\mathrm{int}}
=
-\frac{1}{N_s}
\sum_{i=1}^{N_s}
\sum_{c \in \mathcal{M}}
y_{i,c} \log \pi_{i,c},
\end{equation}
where $\mathcal{M}=\{\mathrm{LK},\mathrm{LLC},\mathrm{RLC}\}$ is the maneuver label set, $y_{i,c}$ is the ground-truth indicator for class $c$, and $\pi_{i,c}$ is the predicted probability for that class.

The trajectory prediction loss is defined as the mean squared error over all retained vehicles and all prediction steps:
\begin{equation}
L_{\mathrm{traj}}
=
\frac{1}{N_s F_{\mathrm{pred}}}
\sum_{i=1}^{N_s}
\sum_{m=1}^{F_{\mathrm{pred}}}
\left\|
\hat{\mathbf{Y}}_i^{m} - \mathbf{Y}_i^{m}
\right\|_2^2,
\end{equation}
where $\hat{\mathbf{Y}}_i^{m}$ and $\mathbf{Y}_i^{m}$ denote the predicted and ground-truth longitudinal--lateral displacement vectors at prediction step $m$.

% To further align maneuver prediction with motion generation, an intention-trajectory consistency term is included. This term penalizes disagreement between the predicted maneuver distribution and the lateral direction implied by the predicted trajectory. Let $\Delta \hat{y}_i$ denote the net predicted lateral displacement of vehicle $v_i$ over the forecast horizon. The consistency loss is written as
% \begin{equation}
% \begin{aligned}
% L_{\mathrm{gate}}
% &=
% \frac{1}{N_s}
% \sum_{i=1}^{N_s}
% \Big[
% \pi_i^{\mathrm{LK}} |\Delta \hat{y}_i| \\
% &\quad
% +
% \pi_i^{\mathrm{LLC}} \max(0,-\Delta \hat{y}_i)
% +
% \pi_i^{\mathrm{RLC}} \max(0,\Delta \hat{y}_i)
% \Big].
% \end{aligned}
% \end{equation}
% This encourages lane-keeping predictions to remain near zero in lateral displacement, left-lane-change predictions to produce leftward motion, and right-lane-change predictions to produce rightward motion.

To further align maneuver prediction with motion generation, an
intention--trajectory consistency term is included. This term penalizes
disagreement between the predicted maneuver distribution and the lateral
direction implied by the predicted trajectory. Let
$\Delta\hat{y}_i$ denote the net predicted lateral displacement of vehicle
$v_i$ over the forecast horizon. Under the adopted coordinate convention,
negative lateral displacement represents leftward motion, while positive
lateral displacement represents rightward motion. The consistency loss is
defined as

\begin{equation}
\begin{aligned}
L_{\mathrm{gate}}
=
\frac{1}{N_s}
\sum_{i=1}^{N_s}
\Big[
&\pi_i^{\mathrm{LK}}\left|\Delta\hat{y}_i\right|
+
\pi_i^{\mathrm{LLC}}
\max\left(0,\Delta\hat{y}_i\right) \\
&+
\pi_i^{\mathrm{RLC}}
\max\left(0,-\Delta\hat{y}_i\right)
\Big].
\end{aligned}
\label{eq:gate_loss}
\end{equation}

This formulation encourages lane-keeping predictions to remain near zero
in lateral displacement, left-lane-change predictions to produce negative
lateral motion, and right-lane-change predictions to produce positive
lateral motion.

The scene-level consistency term introduced earlier is denoted by $L_{\mathrm{scene}}$. Combining all components, the total training objective is
\begin{equation}
L
=
\lambda_{\mathrm{int}} L_{\mathrm{int}}
+
\lambda_{\mathrm{traj}} L_{\mathrm{traj}}
+
\lambda_{\mathrm{gate}} L_{\mathrm{gate}}
+
\lambda_{\mathrm{scene}} L_{\mathrm{scene}},
\end{equation}
where $\lambda_{\mathrm{int}}$, $\lambda_{\mathrm{traj}}$, $\lambda_{\mathrm{gate}}$, and $\lambda_{\mathrm{scene}}$ are nonnegative weights that balance the four terms. In the current design, these weights are set to 10.0, 1.0, 2.0, and 0.5, respectively.  This objective allows the model to learn four properties simultaneously: accurate maneuver classification, accurate future trajectory prediction, agreement between predicted maneuver and predicted motion for each vehicle, and physically compatible futures across the scene.

\section{Experiments}
\label{experiment}

\subsection{Datasets}
The proposed framework is evaluated on two naturalistic driving datasets:
NGSIM and highD. NGSIM provides freeway trajectories from the I-80 and
US-101 study areas at 10 Hz, while highD provides drone-based highway
trajectories that are resampled to 10 Hz for consistency. After trajectory
cleaning, temporal-window extraction, local-scene construction, and
trajectory-group-level splitting, the final scene distributions are
summarized in Table~\ref{tab:dataset_distribution}.

\begin{table*}[!t]
\centering
\caption{Distribution of the final scene-level samples across datasets,
reference-vehicle maneuver classes, and data splits. Each scene is counted
once according to the maneuver of its reference vehicle.}
\label{tab:dataset_distribution}
\renewcommand{\arraystretch}{1.10}
\setlength{\tabcolsep}{7pt}

\begin{tabular}{llrrrr}
\toprule
\textbf{Dataset}
& \textbf{Reference maneuver}
& \textbf{Training}
& \textbf{Validation}
& \textbf{Test}
& \textbf{Total} \\
\midrule

\multirow{4}{*}{NGSIM I-80}
& Lane keeping (LK)       & 5,412 & 1,161 & 1,159 &  7,732 \\
& Left lane change (LLC)  & 2,891 &   621 &   619 &  4,131 \\
& Right lane change (RLC) & 2,523 &   541 &   539 &  3,603 \\
& \textbf{All scenes}     & \textbf{10,826}
                         & \textbf{2,323}
                         & \textbf{2,317}
                         & \textbf{15,466} \\
\midrule

\multirow{4}{*}{NGSIM US-101}
& Lane keeping (LK)       & 3,689 &   792 &   790 &  5,271 \\
& Left lane change (LLC)  & 2,134 &   458 &   457 &  3,049 \\
& Right lane change (RLC) & 1,643 &   352 &   351 &  2,346 \\
& \textbf{All scenes}     & \textbf{7,466}
                         & \textbf{1,602}
                         & \textbf{1,598}
                         & \textbf{10,666} \\
\midrule

\multirow{4}{*}{highD}
& Lane keeping (LK)       &  6,238 & 1,339 & 1,337 &  8,914 \\
& Left lane change (LLC)  &  5,712 & 1,226 & 1,224 &  8,162 \\
& Right lane change (RLC) &  5,050 & 1,084 & 1,082 &  7,216 \\
& \textbf{All scenes}     & \textbf{17,000}
                         & \textbf{3,649}
                         & \textbf{3,643}
                         & \textbf{24,292} \\
\bottomrule
\end{tabular}
\end{table*}

The final scene distributions reflect the different frequencies of
lane-keeping and lane-changing behavior in the three study sites. Each
scene is counted once according to the maneuver of its reference vehicle,
although intention and trajectory targets are generated for every valid
vehicle in the scene. The data are divided into training, validation, and
test subsets using approximate 70\%/15\%/15\% ratio. All temporal windows generated from the same physical trajectory group are assigned to the same subset to prevent vehicle-level and temporal leakage. Since overlapping windows may contain the same physical vehicle more than once, each valid vehicle occurrence is treated as a separate prediction instance within its assigned subset. This preserves the temporal diversity of the extracted scenes while ensuring that observations from the same physical trajectory do not appear across the training, validation, and test sets.

\subsection{Experimental Setup}
For all experiments, each sample is defined using a 7\,s temporal window consisting of a 3\,s observation interval and a 4\,s prediction interval. This corresponds to 30 observation frames and 40 prediction frames after temporal alignment. Scene samples are constructed as described in the previous section, and each scene contains up to 6 vehicles together with a binary validity mask for padded entries. Optimization is performed using AdamW with an initial learning rate of $3\times10^{-4}$, weight decay of $10^{-4}$, gradient clipping at 1.0, and dropout of 0.15. The learning rate is decayed using a cosine annealing schedule over 100 epochs. Training is conducted with a batch size of 8 scene samples.  The interaction and scene-level safety parameters were fixed across all
datasets and experiments. The TTC computation used
$\epsilon=10^{-3}$ to prevent division by zero, while non-closing vehicle
pairs were assigned the saturation value $T_{\max}=10\,\mathrm{s}$.
The scene-level consistency loss and IACR evaluation used a minimum
separation threshold of $D_{\mathrm{safe}}=2.0\,\mathrm{m}$. These values
were held constant during training, validation, and testing.

\subsubsection{Baseline Evaluation Protocol}

All comparison methods were evaluated under the same dataset splits,
observation horizon, prediction horizon, and evaluation samples used for
DSiGAT. Whenever official implementations were available, the baseline
models were retrained using the authors' recommended settings and the
same trajectory-group-level training, validation, and test partitions
adopted in this study. The observation and prediction horizons were fixed
to $3\,\mathrm{s}$ and $4\,\mathrm{s}$, respectively, for all methods.

For methods whose original input representation differed from DSiGAT,
the original architecture-specific feature construction was preserved
whenever required for faithful implementation, while the underlying
trajectory records and evaluation splits remained unchanged. For methods
without an available implementation, the model was reproduced from the
published description and tuned only on the validation split. No test-set
information was used for model selection or hyperparameter adjustment.
All reported intention and trajectory metrics were recomputed on the same
held-out test scenes to ensure a consistent comparison.

\subsection{Evaluation Metrics}

The proposed framework is evaluated from two complementary perspectives. The first measures per-vehicle prediction quality so that the model can be compared directly with prior intention and trajectory prediction methods. The second evaluates scene-level consistency, which is central to the proposed formulation and is not captured by standard single-vehicle metrics alone.

\subsubsection{Per-vehicle intention metrics}

For lane-change intention prediction, standard classification metrics are reported, including accuracy, precision, recall, and macro F1-score. These metrics are computed over the three maneuver classes: lane keeping (LK), left lane change (LLC), and right lane change (RLC). Accuracy measures the overall proportion of correctly classified vehicles, while macro F1-score gives equal weight to all classes and is therefore more informative under class imbalance. Class-wise precision and recall are also reported to show how well the model distinguishes among the three maneuver types. All metrics are computed by pooling predictions from every valid vehicle occurrence across the test scenes, while padded entries are excluded through the validity mask.

To evaluate early anticipation performance, intention prediction is assessed at multiple anticipation times, $T \in \{3\,\mathrm{s}, 2\,\mathrm{s}, 1\,\mathrm{s}, 0\,\mathrm{s}\}$,
where $T=3\,\mathrm{s}$ corresponds to the earliest prediction setting and $T=0$ corresponds to prediction immediately before the annotated lane-change onset, with the onset frame excluded from the input.

\subsubsection{Per-vehicle trajectory metrics}

For trajectory prediction, Root Mean Squared Error (RMSE) is used as the primary displacement-based metric. Let $\hat{\mathbf{y}}_i^m \in \mathbb{R}^{2}$ and
$\mathbf{y}_i^m \in \mathbb{R}^{2}$ denote the predicted and
ground-truth longitudinal--lateral displacement vectors of
vehicle $v_i$ at future step $m$, respectively. The stepwise displacement error is defined as
\begin{equation}
e_i^m = \left\| \hat{\mathbf{y}}_i^m - \mathbf{y}_i^m \right\|_2.
\end{equation}

Using this notation, RMSE is computed over all retained vehicles and prediction steps as

% ADE is computed as the average displacement error over all retained vehicles and all prediction steps,
% \begin{equation}
% \mathrm{ADE}
% =
% \frac{1}{N_s F_{\mathrm{pred}}}
% \sum_{i=1}^{N_s}
% \sum_{m=1}^{F_{\mathrm{pred}}}
% e_i^m.
% \end{equation}

% FDE measures the displacement error at the final prediction step only,
% \begin{equation}
% \mathrm{FDE}
% =
% \frac{1}{N_s}
% \sum_{i=1}^{N_s}
% e_i^{F_{\mathrm{pred}}}.
% \end{equation}

\begin{equation}
\mathrm{RMSE}
=
\sqrt{
\frac{1}{N_s F_{\mathrm{pred}}}
\sum_{i=1}^{N_s}
\sum_{m=1}^{F_{\mathrm{pred}}}
\left(e_i^m\right)^2 }.
\end{equation}

RMSE is reported at prediction horizons of 1\,s, 2\,s, 3\,s, and 4\,s to show how forecasting error evolves as the prediction horizon increases.

\subsubsection{Scene-level metrics}

Standard per-vehicle metrics do not indicate whether the predicted futures
of multiple vehicles remain compatible when considered jointly as one
traffic scene. To evaluate this aspect, two complementary scene-level
metrics are used: Inter-Agent Collision Rate (IACR) and Joint Displacement
Error (JDE). 

The first metric is the Inter-Agent Collision Rate (IACR), defined as the
fraction of predicted scenes in which at least one pair of valid vehicles
violates the minimum separation threshold at any future prediction step.
For scene $s$, the collision indicator is defined as
\begin{equation}
C_s =
\mathbb{I}
\left[
\min_{\substack{i<j\\
m\in\{1,\ldots,F_{\mathrm{pred}}\}}}
\left\|
\hat{\mathbf{p}}_{s,i}^{m}
-
\hat{\mathbf{p}}_{s,j}^{m}
\right\|_2
<
D_{\mathrm{safe}}
\right],
\label{eq:collision_indicator}
\end{equation}
where $D_{\mathrm{safe}}$ is the minimum allowable inter-vehicle
separation. IACR is then computed as
\begin{equation}
\mathrm{IACR}
=
\frac{1}{S}
\sum_{s=1}^{S} C_s,
\end{equation}
where $S$ is the number of evaluated scenes. Lower IACR indicates better
scene-level physical compatibility.

The second metric is the Joint Displacement Error (JDE), which measures
how accurately the predicted trajectories preserve the relative spatial
configuration of all vehicles in a scene. Unlike per-vehicle RMSE, which
evaluates each trajectory independently, JDE compares the predicted and
ground-truth relative displacement between every valid vehicle pair at
each future time step.

For scene $s$, let $\hat{\mathbf{p}}_{s,i}^{m}$ and
$\mathbf{p}_{s,i}^{m}$ denote the predicted and ground-truth
scene-coordinate positions of vehicle $i$ at prediction step $m$,
respectively.
\begin{equation}
d_{s,ij}^{m}
=
\left\|
\left(
\hat{\mathbf{p}}_{s,i}^{m}
-
\hat{\mathbf{p}}_{s,j}^{m}
\right)
-
\left(
\mathbf{p}_{s,i}^{m}
-
\mathbf{p}_{s,j}^{m}
\right)
\right\|_2 .
\label{eq:pairwise_relative_error}
\end{equation}

The Joint Displacement Error is then computed as
\begin{equation}
\mathrm{JDE}
=
\frac{1}{S}
\sum_{s=1}^{S}
\frac{1}{F_{\mathrm{pred}}}
\sum_{m=1}^{F_{\mathrm{pred}}}
\frac{2}{N_s(N_s-1)}
\sum_{i=1}^{N_s-1}
\sum_{j=i+1}^{N_s}
d_{s,ij}^{m},
\label{eq:jde}
\end{equation}
where $S$ is the number of evaluated scenes, $N_s$ is the number of
valid vehicles in scene $s$, and $F_{\mathrm{pred}}$ is the prediction
horizon. Lower JDE indicates that the predicted trajectories better
preserve the joint spatial arrangement and relative motion of the
vehicles in the scene.

\subsection{Implementation Details}

DSiGAT is implemented in PyTorch with PyTorch Geometric for graph construction and graph attention operations. The model is trained on an NVIDIA A100 GPU. Mini-batches contain 8 scene samples, and padded batching is handled with binary validity masks so that dummy nodes do not participate in message passing or loss computation. The node embedding layer maps the raw node features to a 128-dimensional latent space, while the edge embedding layer maps the raw edge features to 64 dimensions. The graph attention module uses 8 attention heads and 3 message-passing rounds at each observation frame. Temporal aggregation is performed with a learned attention mechanism over the 30 observation frames. The intention head is implemented as a two-layer feedforward network with ReLU activation, layer normalization, and dropout. The trajectory module contains three maneuver-specific two-layer feedforward decoders corresponding to lane keeping, left lane change, and right lane change, each producing a 40-step sequence of longitudinal and lateral displacements. 

\section{Results}
\label{results}
This section presents a comprehensive evaluation of the proposed DSiGAT framework. We first assess per-vehicle lane-change intention prediction and trajectory forecasting on the NGSIM I-80 and US-101 datasets, including performance at different anticipation and prediction horizons. We then evaluate generalization on the highD dataset and examine the joint consistency of the predicted multi-vehicle scene using scene-level metrics. Finally, ablation, qualitative, robustness, sensitivity, and computational-efficiency analyses are provided to clarify the contribution, reliability, and practical cost of the proposed framework.

\subsection{Overall Per-Vehicle Intention Prediction Performance}

% Per-vehicle intention prediction is first evaluated to allow direct comparison with existing lane-change prediction methods. In this setting, each vehicle is classified as lane keeping (LK), left lane change (LLC), or right lane change (RLC), while the prediction is still obtained from the proposed scene-level interaction representation. Tables~\ref{tab:intention_i80_overall} and \ref{tab:intention_us101_overall} report the overall results on the NGSIM I-80 and US-101 test sets.

Per-vehicle intention prediction is evaluated over all valid vehicles in the test scenes. Each retained vehicle is classified as lane keeping (LK), left lane change (LLC), or right lane change (RLC), while the prediction is generated from the complete scene-level interaction representation. Metrics are aggregated across all valid vehicle predictions, with padded entries excluded through the validity mask. Tables~\ref{tab:intention_i80_overall} and~\ref{tab:intention_us101_overall} report the resulting performance on the NGSIM I-80 and US-101 test sets.

As shown in Tables~\ref{tab:intention_i80_overall} and \ref{tab:intention_us101_overall}, DSiGAT achieves the best overall accuracy on both datasets, reaching 90.12\% on I-80 and 90.97\% on US-101. Compared with conventional baselines such as SVM, HMM-SVM, CNN-LSTM, IDBN, and LSTM, the proposed model gives clear improvements in both overall accuracy and class-wise F1-score. It also performs better than stronger recent methods such as VWC, Dual Transformer, and MCLG. On I-80, DSiGAT improves the overall accuracy by 2.97\% over MCLG and by 4.15\% over Dual Transformer. On US-101, the corresponding improvements are 2.93\% and 4.00\%. In addition, the proposed model achieves the highest F1-scores for all three maneuver classes on both datasets, reaching 93.45\%, 89.67\%, and 87.23\% for LK, LLC, and RLC on I-80, and 94.12\%, 90.45\%, and 88.34\% on US-101. These results show that the proposed framework improves not only overall recognition accuracy, but also the balance of prediction across different maneuver types.

The precision and recall values further explain this advantage. In lane-change prediction for ADAS, high precision with low recall may cause true lane-change intentions to be missed, while high recall with low precision may lead to unreliable maneuver alarms. This inconsistency appears in several baseline methods. For example, on I-80, Dual Transformer gives the highest RLC precision at 85.55\%, but its recall is lower at 81.45\%. DSiGAT, in contrast, gives 84.43\% precision and 90.22\% recall for the same class, resulting in a stronger F1-score. A similar pattern is observed on US-101, where Dual Transformer records 86.66\% precision and 82.55\% recall for RLC, whereas DSiGAT gives 85.54\% precision and 91.32\% recall. A more consistent balance between precision and recall is also observed for LK and LLC, which indicates that the proposed model provides more dependable maneuver recognition rather than favoring one metric at the expense of the other.

\begin{figure*}
    \centering
    \includegraphics[width=0.7\linewidth]{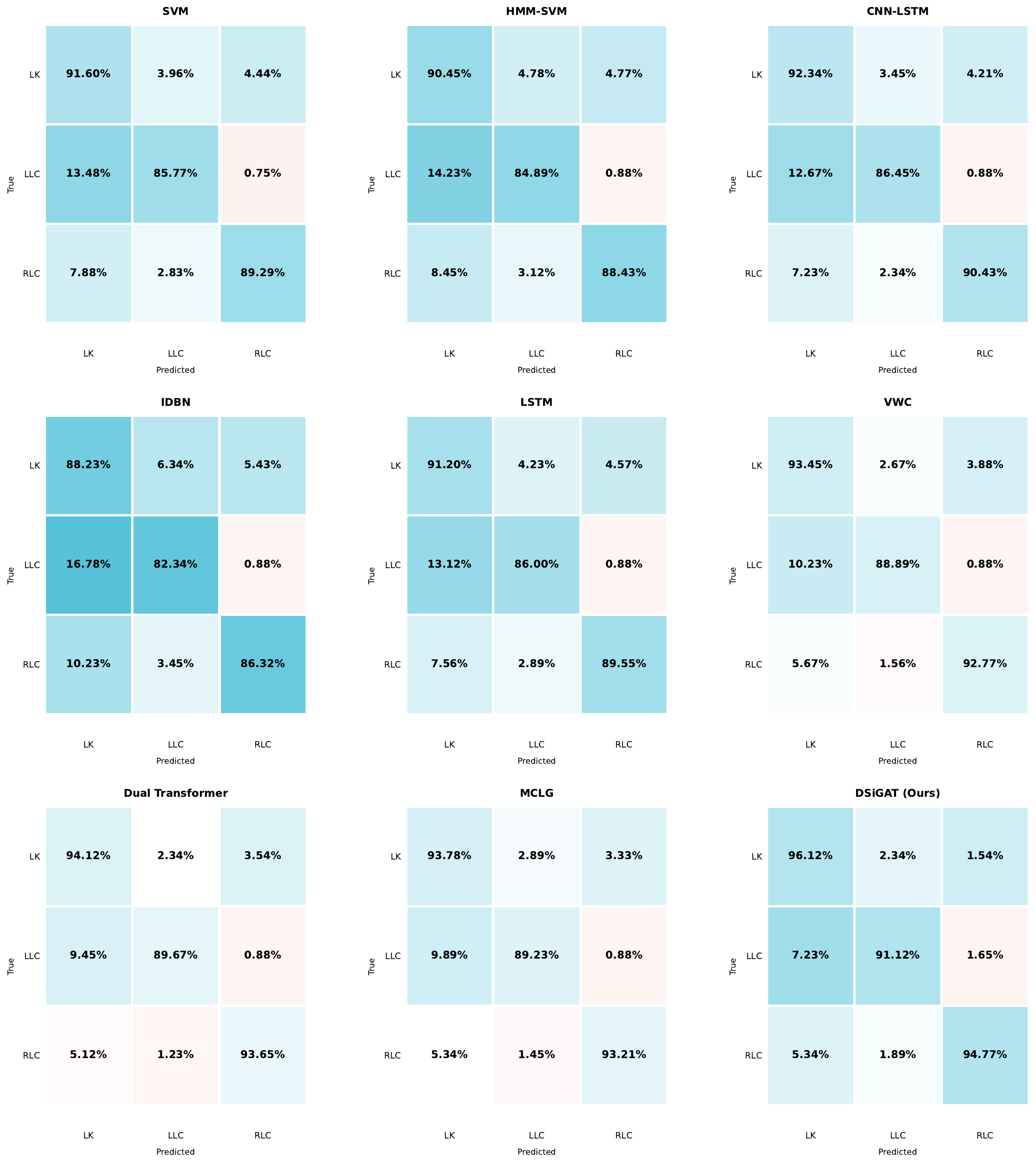}
    \caption{Confusion matrices of different methods for per-vehicle lane-change intention prediction on the NGSIM I-80 test set.}
    \label{fig:confusion_i80}
\end{figure*}

\begin{figure*}
    \centering
    \includegraphics[width=0.7\linewidth]{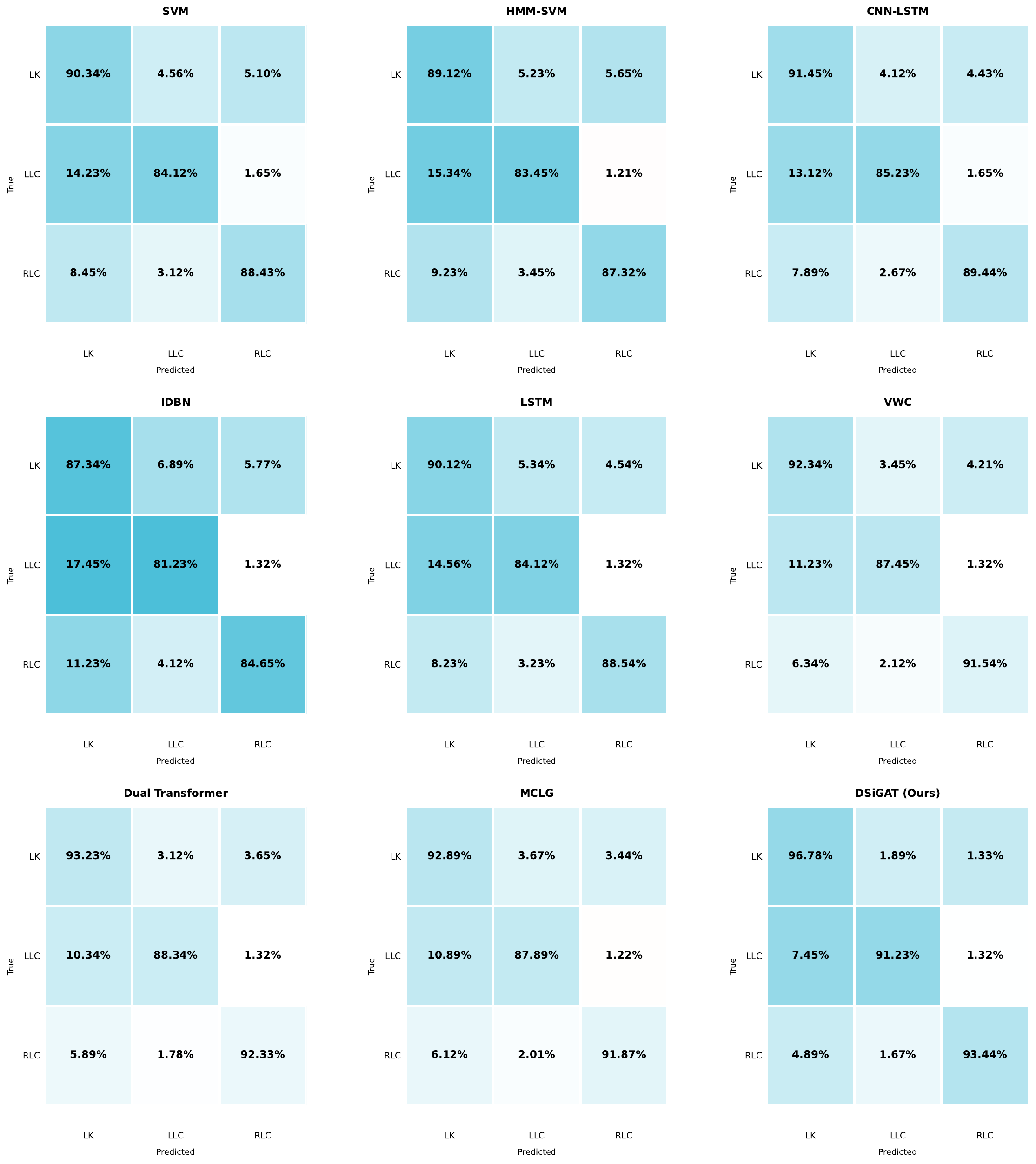}
    \caption{Confusion matrices of different methods for per-vehicle lane-change intention prediction on the NGSIM US-101 test set.}   
    \label{fig:confusion_us101}
\end{figure*}

The confusion matrices in Fig. \ref{fig:confusion_i80} and \ref{fig:confusion_us101} show the same trend. On US-101, DSiGAT correctly classifies 96.78\% of LK samples, 91.23\% of LLC samples, and 93.44\% of RLC samples. On NGSIM I80, the corresponding values are 96.12\%, 91.12\%, and 94.77\%. In both datasets, most of the remaining errors occur when LLC or RLC samples are interpreted as LK, rather than being confused with the opposite lane-change class. This means that the model rarely misunderstands the direction of a lane change, and that the remaining errors mainly appear in transitional cases where the maneuver intent is still forming. Generally, the results indicate that the proposed scene-level interaction modeling leads to stronger and more reliable per-vehicle intention prediction.

\begin{table*}[!htbp]
\centering
\caption{Per-vehicle lane-change intention prediction performance on the NGSIM I-80 test set. Metrics are computed over all valid vehicles across all test scenes; padded entries are excluded. \textcolor{bestred}{\textbf{Red}} = best; \underline{underline} = second best}
\label{tab:intention_i80_overall}
\scriptsize\setlength{\tabcolsep}{4.5pt}\renewcommand{\arraystretch}{1.15}
\begin{adjustbox}{max width=\textwidth}
\begin{tabular}{l c ccc ccc ccc}
\toprule
\multirow{2}{*}{\textbf{Method}} & \multirow{2}{*}{\textbf{Accuracy}}
& \multicolumn{3}{c}{\textbf{LK}}
& \multicolumn{3}{c}{\textbf{LLC}}
& \multicolumn{3}{c}{\textbf{RLC}} \\
\cmidrule(lr){3-5}\cmidrule(lr){6-8}\cmidrule(lr){9-11}
& & P & R & F1 & P & R & F1 & P & R & F1 \\
\midrule
SVM             & 72.75 & 83.65 & 73.85 & 78.45 & 67.13 & 75.86 & 71.23 & 75.36 & 62.88 & 68.56 \\
HMM-SVM         & 71.15 & 70.59 & 84.42 & 76.89 & 72.65 & 66.52 & 69.45 & 62.02 & 73.13 & 67.12 \\
CNN-LSTM         & 76.08 & 85.44 & 77.61 & 81.34 & 79.86 & 69.92 & 74.56 & 76.24 & 68.82 & 72.34 \\
IDBN             & 67.45 & 81.62 & 66.22 & 73.12 & 58.58 & 74.99 & 65.78 & 72.55 & 56.38 & 63.45 \\
LSTM             & 74.52 & 70.03 & 91.20 & 79.23 & 65.65 & 83.35 & 73.45 & 60.59 & 85.40 & 70.89 \\
VWC~\cite{}      & 83.34 & 90.66 & 84.66 & 87.56 & 84.74 & 80.07 & 82.34 & 84.32 & 76.31 & 80.12 \\
Dual Transformer~\cite{} & 85.97 & 91.14 & 87.61 & 89.34 & 86.32 & 83.95 & 85.12 & \textcolor{bestred}{\textbf{85.55}} & 81.45 & 83.45 \\
MCLG~\cite{}     & \underline{87.15} & \underline{88.92} & \underline{91.35} & \underline{90.12} & \underline{84.98} & \underline{88.65} & \underline{86.78} & {83.66} & \underline{85.47} & \underline{84.56} \\
\midrule
\rowcolor{red!5}
\textcolor{bestred}{\textbf{DSiGAT (Ours)}} & \textcolor{bestred}{\textbf{90.12}}
& \textcolor{bestred}{\textbf{92.25}} & \textcolor{bestred}{\textbf{94.68}} & \textcolor{bestred}{\textbf{93.45}}
& \textcolor{bestred}{\textbf{87.37}} & \textcolor{bestred}{\textbf{92.09}} & \textcolor{bestred}{\textbf{89.67}}
& \underline{84.43} & \textcolor{bestred}{\textbf{90.22}} & \textcolor{bestred}{\textbf{87.23}} \\
\bottomrule
\end{tabular}
\end{adjustbox}
\end{table*}

\begin{table*}[!htbp]
\centering
\caption{Per-vehicle lane-change intention prediction performance on the NGSIM US-101 test set. Metrics are computed over all valid vehicles across all test scenes; padded entries are excluded. \textcolor{bestred}{\textbf{Red}} = best; \underline{underline} = second best.}
\label{tab:intention_us101_overall}
\scriptsize\setlength{\tabcolsep}{4.5pt}\renewcommand{\arraystretch}{1.15}
\begin{adjustbox}{max width=\textwidth}
\begin{tabular}{l c ccc ccc ccc}
\toprule
\multirow{2}{*}{\textbf{Method}} & \multirow{2}{*}{\textbf{Accuracy}}
& \multicolumn{3}{c}{\textbf{LK}}
& \multicolumn{3}{c}{\textbf{LLC}}
& \multicolumn{3}{c}{\textbf{RLC}} \\
\cmidrule(lr){3-5}\cmidrule(lr){6-8}\cmidrule(lr){9-11}
& & P & R & F1 & P & R & F1 & P & R & F1 \\
\midrule
SVM             & 73.86 & 84.54 & 74.74 & 79.34 & 68.35 & 77.07 & 72.45 & 76.58 & 64.09 & 69.78 \\
HMM-SVM         & 72.23 & 71.59 & 85.40 & 77.89 & 73.76 & 67.62 & 70.56 & 63.13 & 74.22 & 68.23 \\
CNN-LSTM         & 76.86 & 86.22 & 78.39 & 82.12 & 80.64 & 70.69 & 75.34 & 77.02 & 69.59 & 73.12 \\
IDBN             & 68.56 & 82.73 & 67.31 & 74.23 & 59.69 & 76.06 & 66.89 & 73.66 & 57.46 & 64.56 \\
LSTM             & 75.67 & 71.25 & 92.37 & 80.45 & 66.87 & 84.52 & 74.67 & 61.59 & 86.32 & 71.89 \\
VWC~\cite{}      & 84.34 & 91.44 & 85.44 & 88.34 & 85.85 & 81.18 & 83.45 & 85.43 & 77.42 & 81.23 \\
Dual Transformer~\cite{} & 86.97 & 91.92 & 88.38 & 90.12 & \underline{87.43} & 85.06 & 86.23 & \textcolor{bestred}{\textbf{86.66}} & 82.55 & 84.56 \\
MCLG~\cite{}     & \underline{88.04} & \underline{90.03} & \underline{92.46} & \underline{91.23} & {85.76} & \underline{89.43} & \underline{87.56} & {84.44} & \underline{86.25} & \underline{85.34} \\
\midrule
\rowcolor{red!5}
\textcolor{bestred}{\textbf{DSiGAT (Ours)}} & \textcolor{bestred}{\textbf{90.97}}
& \textcolor{bestred}{\textbf{92.92}} & \textcolor{bestred}{\textbf{95.35}} & \textcolor{bestred}{\textbf{94.12}}
& \textcolor{bestred}{\textbf{88.15}} & \textcolor{bestred}{\textbf{92.87}} & \textcolor{bestred}{\textbf{90.45}}
& \underline{85.54} & \textcolor{bestred}{\textbf{91.32}} & \textcolor{bestred}{\textbf{88.34}} \\
\bottomrule
\end{tabular}
\end{adjustbox}
\end{table*}
 
\subsection{Intention Prediction at Different Anticipation Times}
The next experiment evaluates how early the proposed model can recognize lane-change intention before maneuver onset. For each anticipation time $T\in\{3,2,1,0\}\,\mathrm{s}$, a separate 3-second observation window is constructed so that its final frame occurs $T$ seconds before the annotated lane-change onset; at $T=0$, the onset frame itself is excluded from the input. This analysis is important because reliable early prediction gives an ADAS or autonomous driving system more time to respond to surrounding-vehicle behavior. In this part, F1-score is used as the main metric because it reflects the balance between precision and recall. This is especially important for early intention prediction, where class imbalance and uncertain pre-maneuver behavior can make accuracy alone less informative.

Tables~\ref{tab:intention_i80} and \ref{tab:intention_us101} report the per-class F1-scores at $T=3$\,s, $T=2$\,s, $T=1$\,s, and $T=0$\,s before maneuver onset. As expected, all methods improve as the prediction time approaches $T=0$\,s, since the lane-change cues become more explicit near the actual maneuver point. Even so, DSiGAT maintains the strongest overall performance across the full anticipation range.

In the I-80 dataset, DSiGAT gives the best F1-scores for all three maneuver classes from $T=3$\,s to $T=1$\,s. At $T=3$\,s, it reaches 93.67\%, 89.34\%, and 87.12\% for LK, LLC, and RLC, respectively, which are all higher than the competing methods at the same prediction time. This advantage remains at $T=2$\,s and $T=1$\,s, showing that the proposed model can detect useful maneuver cues even when the lane change is still developing. At $T=0$\,s, DSiGAT still achieves the highest LK score and remains very competitive for LLC and RLC, with only small gaps relative to MCLG.

\begin{figure*}
     \centering
     \includegraphics[width=0.9\linewidth]{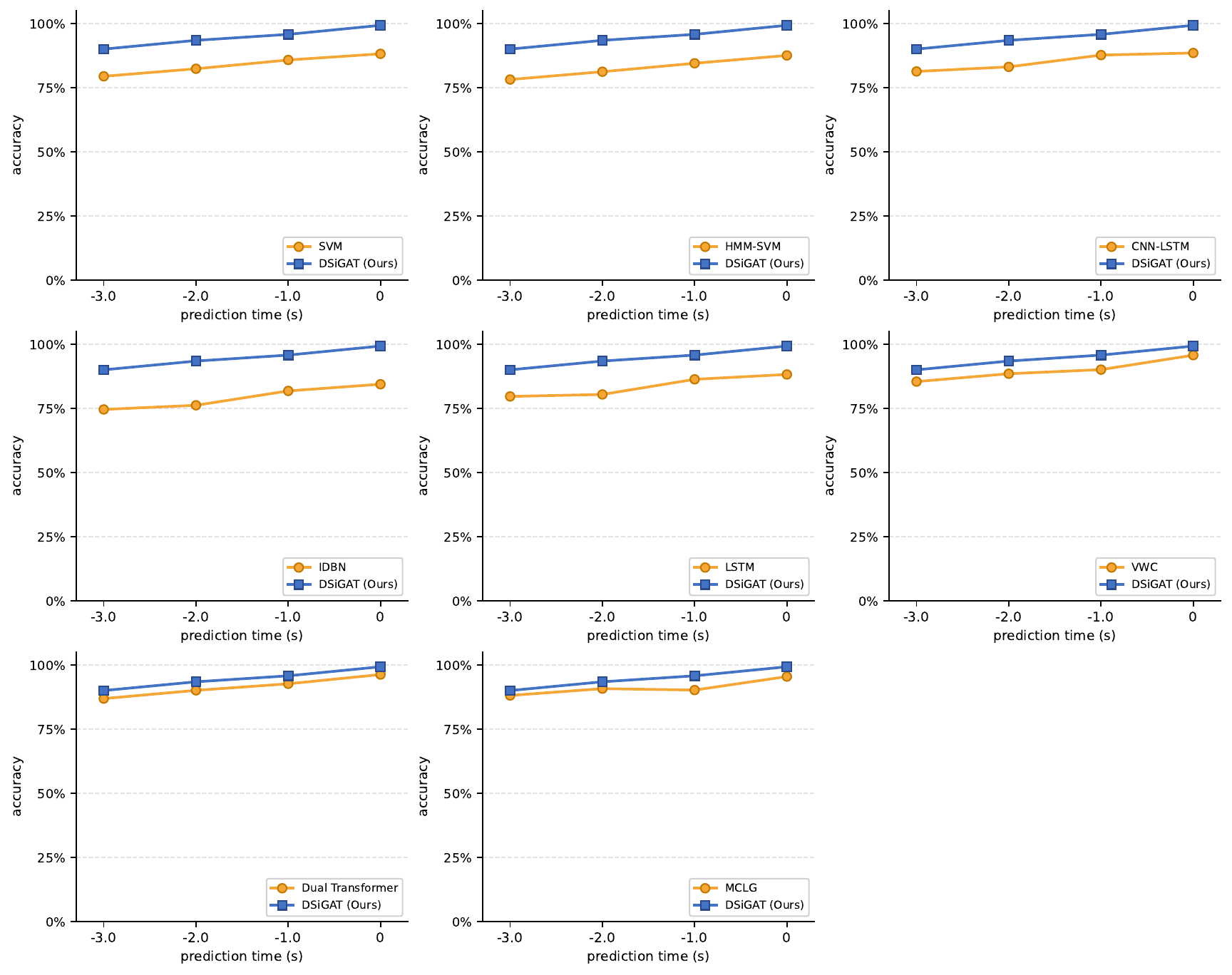}
    \caption{Accuracy comparison of per-vehicle lane-change intention prediction at different anticipation times on the NGSIM I-80 test set.}
    \label{fig:accuracy_i80}
 \end{figure*}

\begin{figure*}
     \centering
     \includegraphics[width=0.9\linewidth]{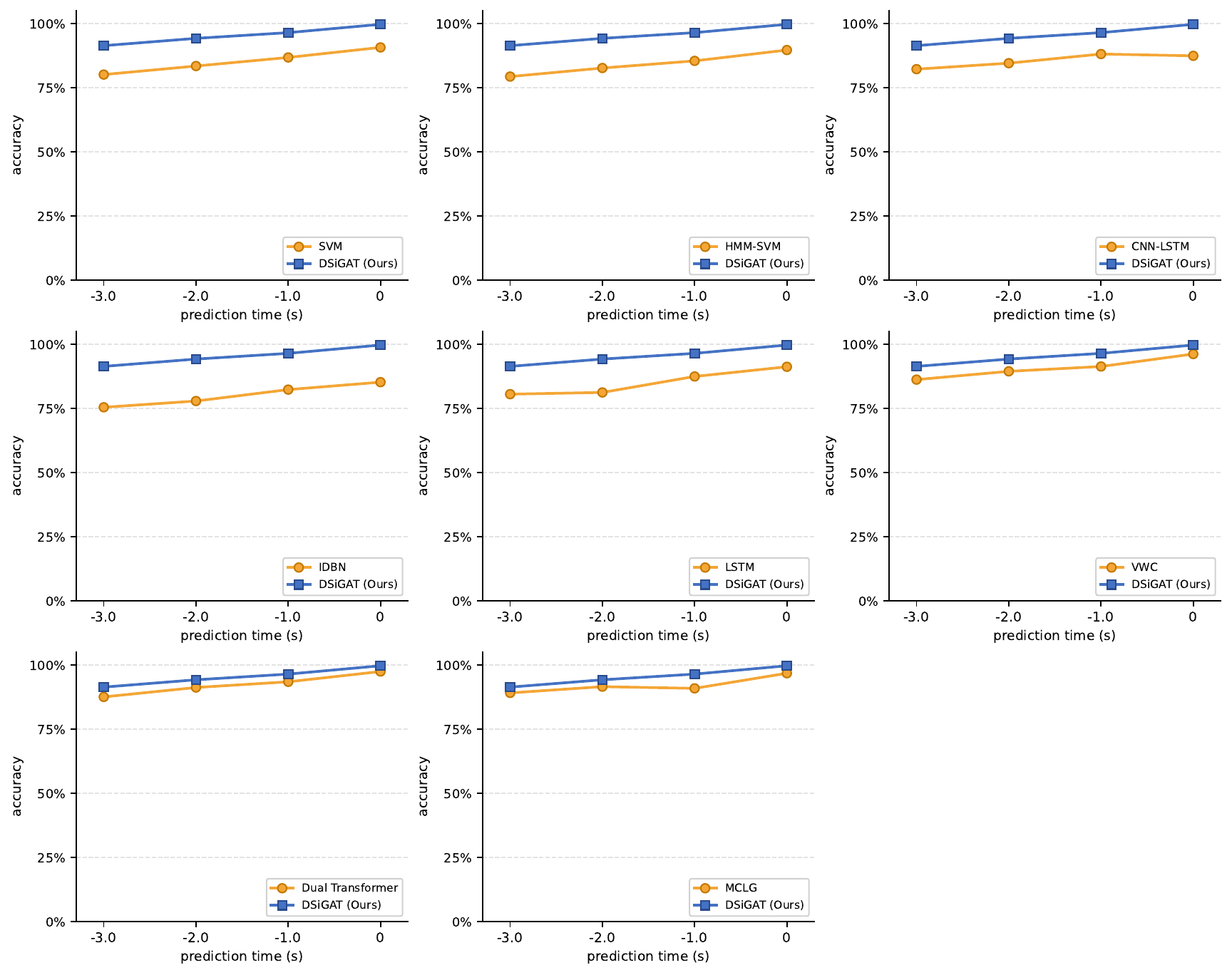}
    \caption{Accuracy of DSiGAT and competing methods at different anticipation times on the NGSIM US-101 test set.}     
    \label{fig:accuracy_us101}
 \end{figure*}

The same trend is observed more clearly on US-101. DSiGAT achieves the best F1-score for LK, LLC, and RLC at every anticipation time. The scores increase from 94.12\%, 92.34\%, and 89.67\% at $T=3$\,s to 99.30\%, 98.61\%, and 96.20\% at $T=0$\,s. These results indicate that the proposed framework performs well not only near maneuver onset, but also at earlier and more difficult prediction times. Fig.~\ref{fig:accuracy_i80} and Fig.~\ref{fig:accuracy_us101} show the same behavior in terms of overall accuracy. In both datasets, DSiGAT remains above each comparison method from $-3$\,s to $0$\,s, and the gap is more visible at earlier prediction times, especially against SVM, HMM-SVM, IDBN, and LSTM-based baselines. The curves also show that the accuracy of all models increases as the maneuver point approaches, but the proposed model maintains a more stable advantage across the full prediction horizon. This suggests that the scene-level graph representation helps DSiGAT capture weak pre-maneuver interaction cues earlier than target-centered baselines. 

% ══════════════════════════════════════════════════════════════
% TABLE 1 — NGSIM I-80
% ══════════════════════════════════════════════════════════════

\begin{table*}[!htbp]
\centering
\caption{Per-class lane-change intention prediction F1-score (\%) on the NGSIM I-80 test set
at different anticipation times. Results are reported for lane keeping (LK), left lane change (LLC),
and right lane change (RLC) at $T=3$\,s, $T=2$\,s, $T=1$\,s, and $T=0$\,s before maneuver onset.
Best results are shown in \textcolor{bestred}{\textbf{red}}, and second-best results are \underline{underlined}.}
\label{tab:intention_i80}
\scriptsize
\setlength{\tabcolsep}{5pt}
\renewcommand{\arraystretch}{1.15}
\begin{adjustbox}{max width=\textwidth}
\begin{tabular}{l ccc ccc ccc ccc}
\toprule
 
\multirow{2}{*}{\textbf{Method}}
& \multicolumn{3}{c}{$T = 3\,\text{s}$}
& \multicolumn{3}{c}{$T = 2\,\text{s}$}
& \multicolumn{3}{c}{$T = 1\,\text{s}$}
& \multicolumn{3}{c}{$T = 0\,\text{s}$} \\
 
\cmidrule(lr){2-4} \cmidrule(lr){5-7}
\cmidrule(lr){8-10} \cmidrule(lr){11-13}
 
& LK & LLC & RLC
& LK & LLC & RLC
& LK & LLC & RLC
& LK & LLC & RLC \\
 
\midrule
 
SVM
& 83.76 & 78.42 & 76.18
& 86.49 & 81.50 & 79.14
& 89.79 & 84.89 & 82.71
& 95.50 & 85.80 & 83.30 \\
 
HMM-SVM
& 82.34 & 77.23 & 75.01
& 85.12 & 80.34 & 78.23
& 88.67 & 83.56 & 81.34
& 94.90 & 84.10 & 83.70 \\
 
CNN-LSTM
& 85.23 & 80.45 & 78.34
& 88.12 & 83.67 & 81.56
& 91.45 & 86.89 & 84.78
& 96.90 & 84.70 & 84.00 \\
 
IDBN
& 79.12 & 73.45 & 71.23
& 82.45 & 76.78 & 74.56
& 86.34 & 80.67 & 78.45
& 92.30 & 81.20 & 79.80 \\
 
LSTM
& 83.45 & 79.34 & 77.23
& 86.23 & 82.56 & 80.45
& 89.56 & 85.78 & 83.67
& 94.40 & 85.80 & 84.50 \\
 
VWC~\cite{}
& 89.34 & 84.56 & 82.45
& 92.12 & 87.78 & 85.67
& 94.89 & 91.01 & 88.89
& \underline{97.44} & 95.56 & 93.99 \\
 
Dual Transformer~\cite{}
& 90.12 & 86.34 & 84.23
& 92.89 & 89.56 & 87.45
& \underline{95.67} & \underline{92.79} & \underline{90.67}
& 96.44 & {94.55} & {93.82} \\
 
MCLG~\cite{}
& \underline{91.45} & \underline{87.56} & \underline{85.34}
& \underline{93.78} & \underline{90.78} & \underline{88.56}
& 95.12 & 91.01 & 88.78
& 96.05 & \textcolor{bestred}{\textbf{95.89}} & \textcolor{bestred}{\textbf{94.50}} \\
 
\midrule
 
\rowcolor{red!5}
\textcolor{bestred}{\textbf{DSiGAT (Ours)}}
& \textcolor{bestred}{\textbf{93.67}}
& \textcolor{bestred}{\textbf{89.34}}
& \textcolor{bestred}{\textbf{87.12}}
& \textcolor{bestred}{\textbf{95.34}}
& \textcolor{bestred}{\textbf{91.78}}
& \textcolor{bestred}{\textbf{89.45}}
& \textcolor{bestred}{\textbf{96.89}}
& \textcolor{bestred}{\textbf{93.67}}
& \textcolor{bestred}{\textbf{91.34}}
& \textcolor{bestred}{\textbf{98.30}}
& \underline{95.62}
& \underline{94.27} \\

\bottomrule
\end{tabular}
\end{adjustbox}
\end{table*}

% ══════════════════════════════════════════════════════════════
% TABLE 2 — NGSIM US-101
% ══════════════════════════════════════════════════════════════
 
\begin{table*}[!htbp]
\centering
\caption{Per-class lane-change intention prediction F1-score (\%) on the NGSIM US-101 test set
at different anticipation times. Results are reported for lane keeping (LK), left lane change (LLC),
and right lane change (RLC) at $T=3$\,s, $T=2$\,s, $T=1$\,s, and $T=0$\,s before maneuver onset.
Best results are shown in \textcolor{bestred}{\textbf{red}}, and second-best results are \underline{underlined}.}
\label{tab:intention_us101}
\scriptsize
\setlength{\tabcolsep}{5pt}
\renewcommand{\arraystretch}{1.15}
\begin{adjustbox}{max width=\textwidth}
\begin{tabular}{l ccc ccc ccc ccc}
\toprule
 
\multirow{2}{*}{\textbf{Method}}
& \multicolumn{3}{c}{$T = 3\,\text{s}$}
& \multicolumn{3}{c}{$T = 2\,\text{s}$}
& \multicolumn{3}{c}{$T = 1\,\text{s}$}
& \multicolumn{3}{c}{$T = 0\,\text{s}$} \\
 
\cmidrule(lr){2-4} \cmidrule(lr){5-7}
\cmidrule(lr){8-10} \cmidrule(lr){11-13}
 
& LK & LLC & RLC
& LK & LLC & RLC
& LK & LLC & RLC
& LK & LLC & RLC \\
 
\midrule
 
SVM
& 84.23 & 79.56 & 77.34
& 87.45 & 82.78 & 80.56
& 90.78 & 85.23 & 83.45
& 96.50 & 88.70 & 86.90 \\
 
HMM-SVM
& 85.34 & 82.12 & 80.23
& 88.56 & 85.34 & 83.45
& 91.89 & 88.67 & 86.78
& 97.60 & 91.80 & 89.60 \\
 
CNN-LSTM
& 83.12 & 74.89 & 72.67
& 86.34 & 78.12 & 75.89
& 89.67 & 81.34 & 79.12
& 95.70 & 84.00 & 82.50 \\
 
IDBN
& 78.45 & 72.34 & 70.12
& 81.67 & 75.56 & 73.34
& 85.89 & 79.78 & 77.56
& 91.80 & 80.40 & 78.60 \\
 
LSTM
& 82.34 & 81.23 & 79.45
& 85.56 & 84.45 & 82.67
& 88.89 & 87.67 & 85.89
& 94.70 & 90.50 & 88.90 \\
 
VWC~\cite{}
& 88.67 & 85.34 & 83.12
& 91.89 & 88.56 & 86.34
& 94.23 & 91.89 & 89.67
& 97.20 & 95.40 & 93.80 \\
 
Dual Transformer~\cite{}
& 90.45 & 87.23 & 85.01
& \underline{93.67} & \underline{90.45} & \underline{88.23}
& \underline{95.89} & \underline{93.12} & \underline{91.34}
& \underline{98.56} & \underline{97.23} & \underline{95.67} \\
 
MCLG~\cite{}
& \underline{91.23} & \underline{88.01} & \underline{85.78}
& 93.45 & 90.23 & 88.01
& 95.67 & 92.89 & 91.12
& 98.34 & 96.78 & 95.12 \\
 
\midrule
 
\rowcolor{red!5}
\textcolor{bestred}{\textbf{DSiGAT (Ours)}}
& \textcolor{bestred}{\textbf{94.12}} & \textcolor{bestred}{\textbf{92.34}} & \textcolor{bestred}{\textbf{89.67}}
& \textcolor{bestred}{\textbf{96.45}} & \textcolor{bestred}{\textbf{95.12}} & \textcolor{bestred}{\textbf{92.89}}
& \textcolor{bestred}{\textbf{97.89}} & \textcolor{bestred}{\textbf{97.23}} & \textcolor{bestred}{\textbf{94.56}}
& \textcolor{bestred}{\textbf{99.30}} & \textcolor{bestred}{\textbf{98.61}} & \textcolor{bestred}{\textbf{96.20}} \\
 
\bottomrule
\end{tabular}
\end{adjustbox}
\end{table*}

\subsection{Per-Vehicle Trajectory Prediction Performance}

To evaluate trajectory prediction performance, the proposed model is compared with a diverse set of ADAS-relevant surrounding-vehicle forecasting baselines from different methodological families, as summarized in Table~\ref{tab:traj_baselines_desc}. This comparison is important because DSiGAT is formulated at the scene level, and its effectiveness should therefore be assessed against methods with different capabilities for modeling interactions among surrounding vehicles.

Tables~\ref{tab:trajectory_i80} and \ref{tab:trajectory_us101} report the trajectory prediction RMSE on the NGSIM I-80 and US-101 test sets over prediction horizons of 1--4\,s. DSiGAT achieves the lowest error at every evaluated horizon on both datasets. Compared with the strongest baseline, GRIP++, the proposed model reduces RMSE on I-80 by 50.00\%, 46.07\%, 38.62\%, and 37.38\% at 1\,s, 2\,s, 3\,s, and 4\,s, respectively. On US-101, the corresponding reductions are 34.15\%, 38.95\%, 34.42\%, and 33.18\%.

Clear differences are also observed across the baseline families. The constant-velocity model produces the largest errors, ranging from 1.23 to 4.98\,m on I-80 and from 1.31 to 5.12\,m on US-101, because it assumes unchanged motion and cannot account for evolving vehicle interactions or maneuver development. Recurrent baselines such as D-LSTM \citep{xin2018intention} and GAIL-GRU \citep{kuefler2017imitating} reduce these errors by learning temporal motion patterns. Social Conv \citep{deo2018convolutional} and MATF GAN \citep{zhao2019multi} further improve prediction by incorporating surrounding-agent influence more explicitly. Graph-based methods achieve lower errors, with GRIP++ providing the strongest baseline results, ranging from 0.38 to 2.14\,m on I-80 and from 0.41 to 2.23\,m on US-101. DSiGAT further reduces these ranges to 0.19--1.34\,m on I-80 and 0.27--1.49\,m on US-101. These results indicate that combining dynamic scene-level interaction reasoning with intention-guided trajectory decoding provides more accurate motion forecasts than trajectory-only formulations.

The advantage of DSiGAT remains consistent as the prediction horizon increases. This is particularly important for ADAS and autonomous driving applications because longer-horizon forecasting depends increasingly on the model's ability to represent evolving scene structure, inter-vehicle influence, and early lane-change preparation. By reasoning jointly over all valid vehicles in the local traffic scene rather than predicting a single target vehicle independently, DSiGAT better preserves relevant interaction information over the full 4-second forecast horizon. The results therefore show that the proposed scene-level design improves not only lane-change intention recognition but also longer-horizon future trajectory prediction.

\begin{table*}[t]
\centering
\caption{Trajectory prediction baseline descriptions.}
\label{tab:traj_baselines_desc}
\renewcommand{\arraystretch}{1.08}
\setlength{\tabcolsep}{6pt}
\begin{tabular}{p{2.3cm} p{10.8cm}}
\toprule
\textbf{Model} & \textbf{Description} \\
\midrule
CV & Constant-velocity kinematic baseline using linear extrapolation from recent motion. \\

D-LSTM \citep{xin2018intention} & Deep recurrent trajectory predictor based on historical motion sequences without explicit scene-graph reasoning. \\

GAIL-GRU \cite{kuefler2017imitating} & GRU-based trajectory predictor trained with adversarial or imitation-style learning to model future motion. \\

Social Conv \cite{deo2018convolutional} & Social interaction model based on convolutional social pooling over neighboring vehicles. \\

MATF GAN \cite{zhao2019multi} & Multi-agent tensor-fusion model that combines interaction and scene context with adversarial multimodal prediction. \\

Dual Transformer \cite{gao2023dual} & Transformer-based spatiotemporal predictor that models temporal evolution and interaction jointly. \\

GRIP \cite{li2019grip} & Graph-based interaction-aware trajectory predictor using graph convolution and recurrent decoding. \\

GRIP++ \cite{li2019grip++} & Enhanced graph-based interaction-aware predictor using both fixed and dynamic graph reasoning. \\

C-VGMM + VIM \cite{deo2018would} & Probabilistic maneuver-aware predictor combining latent motion modes and interaction cues. \\
\bottomrule
\end{tabular}
\end{table*}

\begin{table}[!htbp]
\centering
\caption{Trajectory prediction RMSE (metres, lower is better) on NGSIM I-80.
\textcolor{bestred}{\textbf{Red}} indicates the best result, and
\underline{underlining} indicates the second-best result.}
\label{tab:trajectory_i80}
\scriptsize
\setlength{\tabcolsep}{5pt}
\renewcommand{\arraystretch}{1.15}

\begin{adjustbox}{max width=\columnwidth}
\begin{tabular}{lcccc}
\toprule
\textbf{Method}
& $1\,\text{s}$
& $2\,\text{s}$
& $3\,\text{s}$
& $4\,\text{s}$ \\
\midrule

CV                    & 1.23 & 2.42 & 3.76 & 4.98 \\
D-LSTM                & 0.49 & 1.42 & 2.64 & 4.03 \\
GAIL-GRU              & 0.65 & 1.63 & 2.63 & 3.58 \\
Social Conv           & 0.59 & 1.33 & 2.12 & 3.14 \\
MATF GAN              & 0.61 & 1.29 & 2.04 & 3.12 \\
Dual Transformer      & 1.05 & 2.75 & 4.16 & 5.78 \\
GRIP                   & 0.59 & 1.13 & 1.83 & 2.63 \\
GRIP++                 & \underline{0.38}
                       & \underline{0.89}
                       & \underline{1.45}
                       & \underline{2.14} \\
C-VGMM + VIM          & 0.62 & 1.58 & 2.81 & 4.32 \\

\midrule

\rowcolor{red!5}
\textcolor{bestred}{\textbf{DSiGAT (Ours)}}
& \textcolor{bestred}{\textbf{0.19}}
& \textcolor{bestred}{\textbf{0.48}}
& \textcolor{bestred}{\textbf{0.89}}
& \textcolor{bestred}{\textbf{1.34}} \\

\bottomrule
\end{tabular}
\end{adjustbox}
\end{table}

% ============================================================
% Trajectory Prediction: NGSIM US-101
% ============================================================

\begin{table}[!htbp]
\centering
\caption{Trajectory prediction RMSE (metres, lower is better) on NGSIM US-101.
\textcolor{bestred}{\textbf{Red}} indicates the best result, and
\underline{underlining} indicates the second-best result.}
\label{tab:trajectory_us101}
\scriptsize
\setlength{\tabcolsep}{5pt}
\renewcommand{\arraystretch}{1.15}

\begin{adjustbox}{max width=\columnwidth}
\begin{tabular}{lcccc}
\toprule
\textbf{Method}
& $1\,\text{s}$
& $2\,\text{s}$
& $3\,\text{s}$
& $4\,\text{s}$ \\
\midrule

CV                    & 1.31 & 2.56 & 3.89 & 5.12 \\
D-LSTM                & 0.53 & 1.51 & 2.78 & 4.19 \\
GAIL-GRU              & 0.69 & 1.71 & 2.74 & 3.71 \\
Social Conv           & 0.63 & 1.41 & 2.23 & 3.27 \\
MATF GAN              & 0.65 & 1.37 & 2.15 & 3.24 \\
Dual Transformer      & 1.09 & 2.83 & 4.27 & 5.91 \\
GRIP                   & 0.63 & 1.21 & 1.94 & 2.76 \\
GRIP++                 & \underline{0.41}
                       & \underline{0.95}
                       & \underline{1.54}
                       & \underline{2.23} \\
C-VGMM + VIM          & 0.66 & 1.65 & 2.93 & 4.45 \\

\midrule

\rowcolor{red!5}
\textcolor{bestred}{\textbf{DSiGAT (Ours)}}
& \textcolor{bestred}{\textbf{0.27}}
& \textcolor{bestred}{\textbf{0.58}}
& \textcolor{bestred}{\textbf{1.01}}
& \textcolor{bestred}{\textbf{1.49}} \\

\bottomrule
\end{tabular}
\end{adjustbox}
\end{table}

\newcommand{\cmark}{\checkmark}
\newcommand{\xmark}{$\times$}

\subsection{Ablation Study}

Table~\ref{tab:ablation_dsigat} presents a progressive ablation study of the proposed DSiGAT framework on the NGSIM I-80 and US-101 datasets. The study begins with a simple single-vehicle baseline and then introduces the proposed components one at a time, namely multi-vehicle context (MVC), dynamic scene graph construction (DSG), edge features (EF), temporal graph attention message passing (TGAM), intention-guided decoding (IGD), and the scene-level consistency loss (SCL). This progressive design is important because it shows how each component contributes to maneuver recognition, especially for the more difficult lane-change classes.

Starting from the baseline model (which uses the historical kinematic features of the target), the addition of MVC improves the F1-scores of all classes in both datasets, which confirms that surrounding vehicles already provide useful intention cues even before explicit graph reasoning is introduced. After the dynamic scene graph is added, the gains become much clearer, especially for LLC and RLC. In I-80, the LLC and RLC F1-scores increase from 64.78\% and 60.23\% to 72.45\% and 68.56\%, while in US-101 they increase from 65.89\% and 61.45\% to 73.45\% and 69.78\%. This shows that explicit interaction structure is more effective than simple contextual aggregation.

The addition of edge features further improves the results across all classes, indicating that relational quantities such as gap, lateral offset, relative speed, and TTC provide meaningful geometric cues for lane-change prediction. When temporal graph attention is introduced, the gains become more pronounced again. The mean F1-score rises from 80.17\% to 85.71\%, and the improvement is especially visible in the lane-change classes, which suggests that the model benefits from learning both which surrounding vehicles matter and which observation frames are most informative.

Adding intention-guided decoding provides another consistent gain, increasing the mean F1-score from 85.71\% to 88.60\%. This indicates that explicitly coupling the learned representation with maneuver-specific decoding helps stabilize class discrimination. The final improvement is obtained after introducing the scene-level consistency loss, which raises the mean F1-score to 90.54\% and gives the best class-wise performance on both datasets. Compared with the baseline, the full DSiGAT model improves LK, LLC, and RLC by 22.22\%, 31.33\%, and 33.11\% on I-80, and by 21.67\%, 31.33\%, and 33.00\% on US-101. These results show that each component contributes positively, and that the full scene-level design is particularly beneficial for the more challenging lane-change classes.

\begin{table*}[t]
\centering
\caption{Progressive ablation study of the proposed DSiGAT framework on the NGSIM I-80 and US-101 datasets. Best results are shown in \textbf{bold}.}
\label{tab:ablation_dsigat}
\renewcommand{\arraystretch}{1.08}
\setlength{\tabcolsep}{4pt}
\begin{tabular}{l|cccccc|ccc|ccc|c}
\toprule
\textbf{Model Variant} & \textbf{MVC} & \textbf{DSG} & \textbf{EF} & \textbf{TGAM} & \textbf{IGD} & \textbf{SCL}
& \multicolumn{3}{c|}{\textbf{I-80 F1 (\%)}}
& \multicolumn{3}{c|}{\textbf{US-101 F1 (\%)}}
& \textbf{Mean} \\
\cline{8-13}
& & & & & & & \textbf{LK} & \textbf{LLC} & \textbf{RLC} & \textbf{LK} & \textbf{LLC} & \textbf{RLC} & \textbf{F1} \\
\midrule

Baseline
& $\times$ & $\times$ & $\times$ & $\times$ & $\times$ & $\times$
& 71.23 & 58.34 & 54.12
& 72.45 & 59.12 & 55.34
& 61.77 \\

$+$ MVC
& $\checkmark$ & $\times$ & $\times$ & $\times$ & $\times$ & $\times$
& 76.45 & 64.78 & 60.23
& 77.56 & 65.89 & 61.45
& 67.73 \\

$+$ DSG
& $\checkmark$ & $\checkmark$ & $\times$ & $\times$ & $\times$ & $\times$
& 81.34 & 72.45 & 68.56
& 82.23 & 73.45 & 69.78
& 74.64 \\

$+$ EF
& $\checkmark$ & $\checkmark$ & $\checkmark$ & $\times$ & $\times$ & $\times$
& 85.67 & 78.23 & 75.34
& 86.45 & 79.12 & 76.23
& 80.17 \\

$+$ TGAM
& $\checkmark$ & $\checkmark$ & $\checkmark$ & $\checkmark$ & $\times$ & $\times$
& 89.12 & 84.56 & 81.78
& 90.23 & 85.67 & 82.89
& 85.71 \\

$+$ IGD
& $\checkmark$ & $\checkmark$ & $\checkmark$ & $\checkmark$ & $\checkmark$ & $\times$
& 91.89 & 87.34 & 85.12
& 92.67 & 88.23 & 86.34
& 88.60 \\

\midrule

Full DSiGAT
& $\checkmark$ & $\checkmark$ & $\checkmark$ & $\checkmark$ & $\checkmark$ & $\checkmark$
& \textbf{93.45} & \textbf{89.67} & \textbf{87.23}
& \textbf{94.12} & \textbf{90.45} & \textbf{88.34}
& \textbf{90.54} \\

\bottomrule
\end{tabular}
\vspace{2mm}

\footnotesize
MVC: Multi-Vehicle Context; DSG: Dynamic Scene Graph; EF: Edge Features;TGAM: Temporal Graph Attention Message Passing; IGD: Intention-Guided Decoding;
SCL: Scene-Level Consistency Loss.

\end{table*}

\subsection{Comparison on an Additional Dataset}
In this study, additional experiments are conducted on the highD dataset to examine whether the proposed framework maintains its performance beyond the NGSIM setting. For consistency, the same input feature design used in the NGSIM experiments is retained here, including per-vehicle motion features, multi-vehicle context, and interaction-related descriptors. This allows the comparison to focus on the model itself rather than changes in feature construction.

Table~\ref{tab:intention_highD} reports the per-class lane-change intention prediction F1-scores at different anticipation times. DSiGAT achieves the best results for all maneuver classes across all anticipation times. At $T=3$\,s, the proposed model reaches 94.78\%, 91.23\%, and 89.45\% for LK, LLC, and RLC, respectively. At $T=0$\,s, these values increase to 99.45\%, 97.89\%, and 96.34\%. Compared with the strongest competing method, the gains are 2.44\%, 3.00\%, and 3.00\% at $T=3$\,s, and 1.14\%, 1.14\%, and 0.58\% at $T=0$\,s for LK, LLC, and RLC, respectively.

Table~\ref{tab:trajectory_highD} reports the trajectory prediction RMSE on the highD dataset over prediction horizons of 1--4\,s. DSiGAT achieves the lowest error at 1\,s, 3\,s, and 4\,s, while remaining highly competitive at 2\,s. At this horizon, DSiGAT records an RMSE of 0.79\,m, which is only 0.01\,m higher than the 0.78\,m obtained by GRIP++. Compared with GRIP++, the proposed model reduces RMSE by 52.94\%, 11.38\%, and 29.63\% at 1\,s, 3\,s, and 4\,s, respectively. At 2\,s, DSiGAT exhibits a marginal RMSE increase of 1.28\%. Overall, the results follow the pattern observed on NGSIM and indicate that dynamic scene-level interaction reasoning supports accurate future trajectory prediction across most evaluated horizons.

The highD results demonstrate that the proposed framework remains effective for both lane-change intention prediction and future trajectory forecasting under a highway dataset with different recording conditions and traffic characteristics. Although DSiGAT is marginally outperformed at the 2-second trajectory horizon, it achieves the best performance at the remaining three horizons and maintains strong intention prediction results. These findings indicate that the effectiveness of DSiGAT is not limited to the NGSIM data distribution and that its scene-level interaction modeling generalizes to another large-scale naturalistic driving benchmark.

% Table~\ref{tab:trajectory_highD} reports the trajectory prediction RMSE on the same dataset. DSiGAT again gives the lowest error at all prediction horizons. Compared with GRIP++, the proposed model reduces RMSE by 50.00\%, 19.39\%, 11.38\%, 29.63\%, and 19.92\% at 1\,s through 5\,s, respectively. These results follow the same pattern observed on NGSIM, where stronger scene-level interaction reasoning leads to more accurate future trajectory prediction.

% The highD results show that the proposed framework remains effective for both lane-change intention prediction and future trajectory forecasting under a different highway dataset. This indicates that the gains of DSiGAT are not tied only to the NGSIM data distribution, but remain consistent when evaluated on another large-scale naturalistic driving benchmark.

% ══════════════════════════════════════════════════════════════
% TABLE — Intention Prediction, highD
% ══════════════════════════════════════════════════════════════
 
\begin{table*}[!htbp]
\centering
\caption{Per-class lane-change intention prediction F1-score (\%) on the
highD dataset at different anticipation times $T_p$.
\textcolor{bestred}{\textbf{Red}} = best; \underline{underline} = second best.}
\label{tab:intention_highD}
\scriptsize
\setlength{\tabcolsep}{5pt}
\renewcommand{\arraystretch}{1.15}
\begin{adjustbox}{max width=\textwidth}
\begin{tabular}{l ccc ccc ccc ccc}
\toprule
 
\multirow{2}{*}{\textbf{Method}}
& \multicolumn{3}{c}{$T = 3\,\text{s}$}
& \multicolumn{3}{c}{$T = 2\,\text{s}$}
& \multicolumn{3}{c}{$T = 1\,\text{s}$}
& \multicolumn{3}{c}{$T = 0\,\text{s}$} \\
 
\cmidrule(lr){2-4} \cmidrule(lr){5-7}
\cmidrule(lr){8-10} \cmidrule(lr){11-13}
 
& LK & LLC & RLC
& LK & LLC & RLC
& LK & LLC & RLC
& LK & LLC & RLC \\
 
\midrule
 
SVM
& 85.34 & 80.12 & 78.23
& 88.23 & 83.45 & 81.56
& 91.45 & 86.78 & 84.89
& 97.12 & 89.34 & 87.45 \\
 
HMM-SVM
& 84.12 & 78.89 & 77.01
& 87.01 & 82.23 & 80.34
& 90.23 & 85.45 & 83.56
& 96.34 & 87.89 & 86.01 \\
 
CNN-LSTM
& 86.45 & 81.34 & 79.56
& 89.34 & 84.67 & 82.78
& 92.56 & 87.89 & 85.89
& 97.67 & 86.12 & 85.23 \\
 
IDBN
& 80.23 & 74.56 & 72.78
& 83.45 & 77.89 & 75.89
& 87.12 & 81.23 & 79.34
& 93.34 & 82.34 & 80.56 \\
 
LSTM
& 84.67 & 80.45 & 78.56
& 87.56 & 83.78 & 81.89
& 90.78 & 87.01 & 85.12
& 95.89 & 87.23 & 85.89 \\
 
VWC~\cite{}
& 90.12 & 85.78 & 83.89
& 93.23 & 88.89 & 86.89
& 95.89 & 92.12 & 90.23
& \underline{98.34} & 96.45 & 94.89 \\
 
Dual Transformer~\cite{}
& 91.23 & 87.01 & 85.23
& 93.89 & 90.23 & 88.34
& \underline{96.45} & \underline{93.34} & \underline{91.45}
& 97.89 & \underline{97.12} & \underline{95.78} \\
 
MCLG~\cite{}
& \underline{92.34} & \underline{88.23} & \underline{86.45}
& \underline{94.78} & \underline{91.56} & \underline{89.67}
& 96.12 & 92.89 & 90.78
& 97.45 & 96.78 & 95.34 \\
 
\midrule
 
\rowcolor{red!5}
\textcolor{bestred}{\textbf{DSiGAT (Ours)}}
& \textcolor{bestred}{\textbf{94.78}}
& \textcolor{bestred}{\textbf{91.23}}
& \textcolor{bestred}{\textbf{89.45}}
& \textcolor{bestred}{\textbf{96.89}}
& \textcolor{bestred}{\textbf{93.67}}
& \textcolor{bestred}{\textbf{91.78}}
& \textcolor{bestred}{\textbf{98.12}}
& \textcolor{bestred}{\textbf{95.45}}
& \textcolor{bestred}{\textbf{93.56}}
& \textcolor{bestred}{\textbf{99.45}}
& \textcolor{bestred}{\textbf{97.89}}
& \textcolor{bestred}{\textbf{96.34}} \\
 
\bottomrule
\end{tabular}
\end{adjustbox}
\end{table*}
 
% ══════════════════════════════════════════════════════════════
% TABLE — Trajectory Prediction RMSE, highD
% ══════════════════════════════════════════════════════════════
 
% \begin{table}[!htbp]
% \centering
% \caption{Trajectory prediction RMSE (metres, lower is better) on highD.
% \textcolor{bestred}{\textbf{Red}} = best; \underline{underline} = second best.}
% \label{tab:trajectory_highD}
% \scriptsize
% \setlength{\tabcolsep}{5pt}
% \renewcommand{\arraystretch}{1.15}
% \begin{adjustbox}{max width=\columnwidth}
% \begin{tabular}{l ccccc}
% \toprule
 
% \textbf{Method}
% & $1\,\text{s}$
% & $2\,\text{s}$
% & $3\,\text{s}$
% & $4\,\text{s}$
% & $5\,\text{s}$ \\
 
% \midrule
 
% CV
% & 1.12 & 2.23 & 3.45 & 4.67 & 5.89 \\
 
% D-LSTM
% & 0.45 & 1.23 & 2.34 & 3.56 & 4.89 \\
 
% GAIL-GRU
% & 0.56 & 1.34 & 2.23 & 3.12 & 4.23 \\
 
% Social Conv
% & 0.48 & 1.12 & 1.89 & 2.78 & 3.89 \\
 
% MATF GAN
% & 0.52 & 1.18 & 1.95 & 2.89 & 3.78 \\
 
% Dual Transformer~\cite{}
% & 0.89 & 2.34 & 3.56 & 4.89 & 5.67 \\
 
% GRIP
% & 0.45 & 0.98 & 1.56 & 2.23 & 2.78 \\
 
% GRIP++
% & \underline{0.34} & \textcolor{bestred}{\textbf{0.78}} & \underline{1.23} & \underline{1.89} & \underline{2.56} \\
 
% C-VGMM + VIM
% & 0.56 & 1.34 & 2.45 & 3.67 & 5.12 \\
 
% \midrule
 
% \rowcolor{red!5}
% \textcolor{bestred}{\textbf{DSiGAT (Ours)}}
% & \textcolor{bestred}{\textbf{0.16}}
% & \underline{0.79}
% & \textcolor{bestred}{\textbf{1.09}}
% & \textcolor{bestred}{\textbf{1.33}}
% & \textcolor{bestred}{\textbf{2.05}} \\
 
% \bottomrule
% \end{tabular}
% \end{adjustbox}
% \end{table}

\begin{table}[!htbp]
\centering
\caption{Trajectory prediction RMSE (metres, lower is better) on highD.
\textcolor{bestred}{\textbf{Red}} indicates the best result, and
\underline{underlining} indicates the second-best result.}
\label{tab:trajectory_highD}
\scriptsize
\setlength{\tabcolsep}{5pt}
\renewcommand{\arraystretch}{1.15}

\begin{adjustbox}{max width=\columnwidth}
\begin{tabular}{lcccc}
\toprule

\textbf{Method}
& $1\,\text{s}$
& $2\,\text{s}$
& $3\,\text{s}$
& $4\,\text{s}$ \\

\midrule

CV
& 1.12 & 2.23 & 3.45 & 4.67 \\

D-LSTM
& 0.45 & 1.23 & 2.34 & 3.56 \\

GAIL-GRU
& 0.56 & 1.34 & 2.23 & 3.12 \\

Social Conv
& 0.48 & 1.12 & 1.89 & 2.78 \\

MATF GAN
& 0.52 & 1.18 & 1.95 & 2.89 \\

Dual Transformer
& 0.89 & 2.34 & 3.56 & 4.89 \\

GRIP
& 0.45 & 0.98 & 1.56 & 2.23 \\

GRIP++
& \underline{0.34}
& \textcolor{bestred}{\textbf{0.78}}
& \underline{1.23}
& \underline{1.89} \\

C-VGMM + VIM
& 0.56 & 1.34 & 2.45 & 3.67 \\

\midrule

\rowcolor{red!5}
\textcolor{bestred}{\textbf{DSiGAT (Ours)}}
& \textcolor{bestred}{\textbf{0.16}}
& \underline{0.79}
& \textcolor{bestred}{\textbf{1.09}}
& \textcolor{bestred}{\textbf{1.33}} \\

\bottomrule
\end{tabular}
\end{adjustbox}
\end{table}

\subsection{Qualitative Analysis}
\subsubsection{Representative Successful Cases}
\begin{figure*}
    \centering
    \includegraphics[width=1\linewidth]{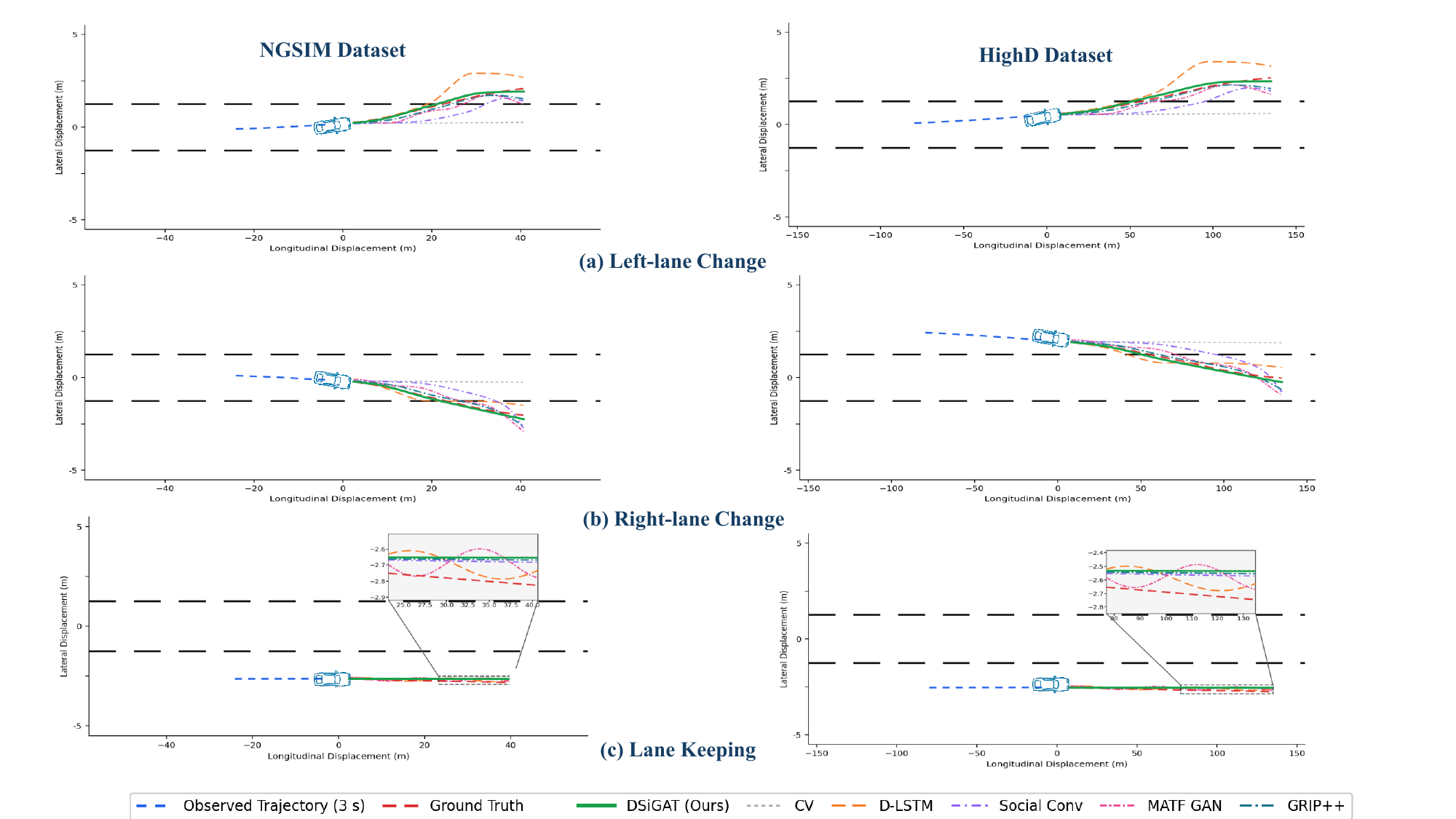}
    \caption{Qualitative comparison of lane-change trajectory prediction under scene context for representative left-lane-change and right-lane-change cases on the NGSIM and highD datasets against competitive baselines.}
    \label{fig:qualitative_baseline}
\end{figure*}

\begin{figure*}
    \centering
    \includegraphics[width=1\linewidth]{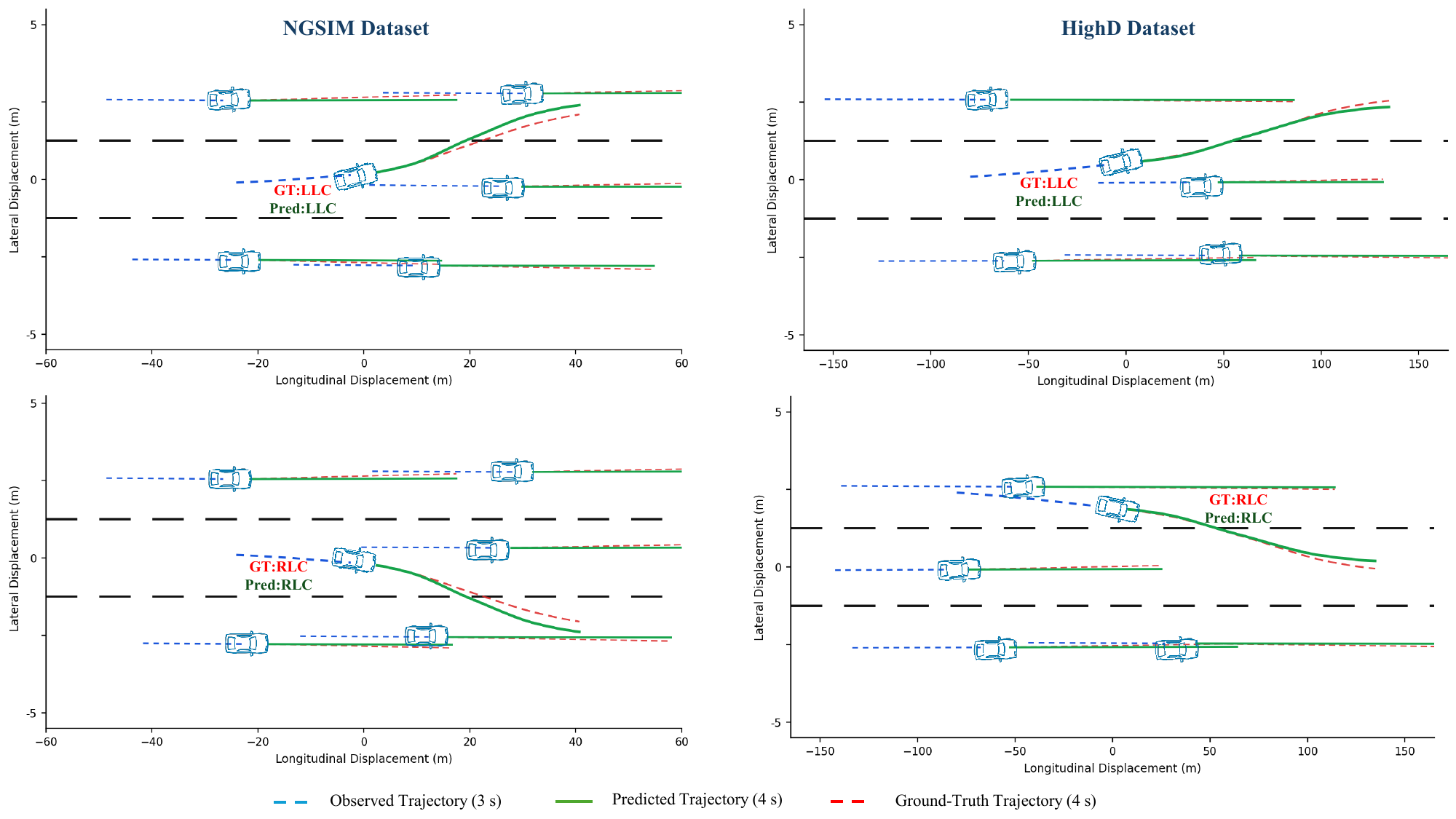}
    \caption{Qualitative examples of joint lane-change intention and trajectory prediction on the NGSIM and highD datasets. Ground-truth and predicted maneuver labels are shown for the key lane-changing vehicle in each scene, while the remaining vehicles in the scene are correctly predicted as lane keeping and therefore are not labeled individually for clarity.}    \label{fig:qualitative_scene}
\end{figure*}

To further examine how the proposed framework behaves under representative highway traffic scenarios, Figs.~\ref{fig:qualitative_baseline} and \ref{fig:qualitative_scene} present qualitative results from the NGSIM and highD datasets. Fig.~\ref{fig:qualitative_baseline} compares DSiGAT with competing trajectory prediction methods for representative left-lane-change, right-lane-change, and lane-keeping cases, while Fig.~\ref{fig:qualitative_scene} presents scene-level examples of the proposed framework together with the ground-truth and predicted maneuver labels of the key lane-changing vehicle.

As shown in Fig.~\ref{fig:qualitative_baseline}, DSiGAT remains closer to the ground-truth future trajectory than the competing baselines across all three maneuver types. This advantage is most visible during the lateral transition phase of lane changes, where several baselines either begin the maneuver too early, respond too late, or deviate more strongly from the actual path. By contrast, the predicted trajectory of DSiGAT follows the ground-truth evolution more closely and preserves a smoother lane transition. This behavior is consistent with the architecture of the proposed model. Since DSiGAT represents the local traffic scene as a dynamic graph, each vehicle is not predicted only from its own motion history, but from the evolving interaction among surrounding vehicles. The edge features encode relative gap, lateral offset, relative speed, and related interaction cues, while temporal graph attention message passing allows the model to focus on the most informative vehicles and time steps before the maneuver occurs. As a result, the model better captures when a lane change is likely to begin and how its lateral motion will develop. 

Fig.~\ref{fig:qualitative_scene} further clarifies the scene-level role of the model. In the displayed NGSIM and highD examples, the predicted trajectory of the key maneuvering vehicle closely matches the actual lane-change path, and the predicted maneuver label agrees with the ground truth for both LLC and RLC cases. At the same time, the surrounding vehicles preserve behavior that is compatible with the evolving traffic scene. This is an important point of distinction from target-centered prediction. In DSiGAT, the temporal graph attention mechanism propagates influence among vehicles through the scene graph, and the intention-guided decoder further constrains the predicted future motion to remain consistent with the inferred maneuver class. This is why the resulting trajectory is not only close to the ground truth, but also behaviorally coherent with the surrounding vehicles in the scene.

The lane-keeping examples provide a similar observation. Even when no lane change occurs, DSiGAT generates a more stable future path with less drift than the competing methods. This suggests that the benefit of scene-level reasoning is not limited to active maneuvers, but also helps preserve motion consistency under ordinary lane-following behavior. In sum, the qualitative results show that the proposed architecture improves prediction in two related ways: it captures the timing and shape of lane-change trajectories more accurately, and it does so through explicit scene-level interaction reasoning rather than isolated vehicle modeling.

\subsubsection{Representative Failure Cases}

\begin{figure*}
    \centering
    \includegraphics[width=1\linewidth]{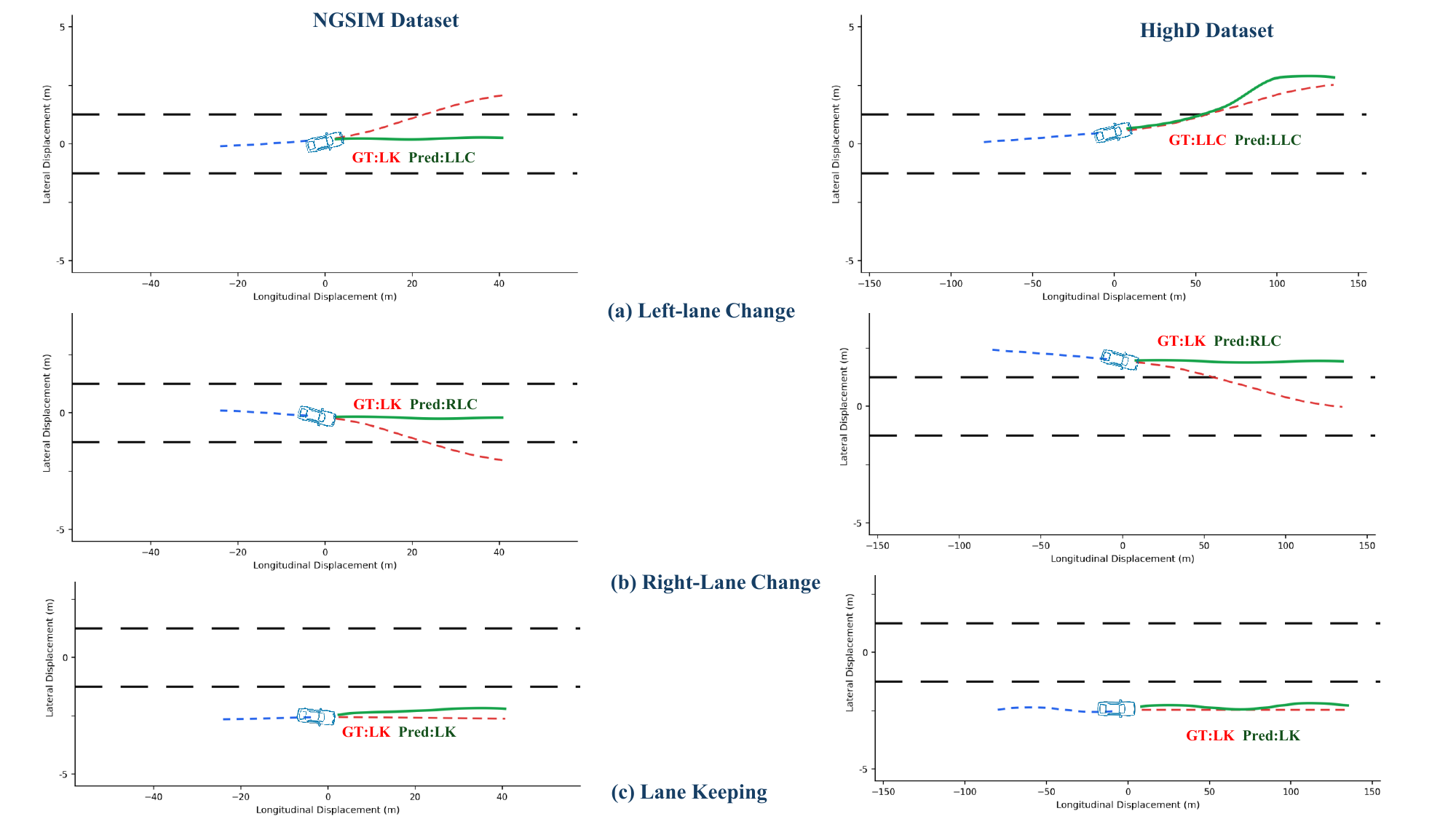}
    \caption{Representative failure cases of DSiGAT on the NGSIM and highD datasets. The examples illustrate situations in which weak or ambiguous early maneuver cues lead to intention misclassification, particularly between lane keeping and lane-change behaviors, while the predicted trajectories remain smooth and physically plausible.}
    \label{fig:failure_cases}
\end{figure*}

Fig.~\ref{fig:failure_cases} presents representative failure cases that reveal the remaining limitations of the proposed framework under ambiguous maneuver conditions. These examples show that the main errors arise in situations where the early motion pattern is weak, the lateral transition remains ambiguous during the observation window, or the vehicle exhibits short-term behavior that resembles the onset of a lane change but does not fully develop into one.

In the NGSIM examples, the failure cases show that DSiGAT may occasionally predict a left- or right-lane change when the ground-truth maneuver remains lane keeping. This behavior suggests that, in some scenes, the model interprets small lateral drift or transient interaction cues as sufficient evidence for a lane-change intention. A similar pattern is also observed in the highD examples, where a lane-keeping trajectory is predicted as a right-lane change. These cases indicate that the most difficult situations are not the fully developed lane changes, but the borderline scenes in which early maneuver cues are present but remain incomplete. These failure patterns are consistent with the nature of highway behavior prediction. Lane-change intention is often formed gradually, and early motion cues may be subtle, noisy, or temporarily misleading. Recent prediction studies \cite{mozaffari2022early} similarly note that ambiguous interaction context, weak maneuver onset, and uncertainty in future lane commitment remain challenging even for strong trajectory and behavior models. 

Even in these failure cases, the predicted trajectories generally remain smooth and physically plausible, which suggests that the scene-level interaction modeling still preserves reasonable motion structure even when the maneuver class is misidentified. Overall, the failure scenarios indicate that the remaining challenge is less about generating unstable trajectories and more about distinguishing weak lane-change intent from ordinary lane-keeping behavior at very early stages.

\subsection{Scene-Level Consistency and Prediction Performance}

Standard per-vehicle metrics do not indicate whether the predicted futures
of multiple vehicles remain compatible when evaluated jointly as one traffic
scene. To assess this aspect, Table~\ref{tab:scene_level_metrics} reports
the scene-level results using Inter-Agent Collision Rate (IACR) and Joint
Displacement Error (JDE). These metrics complement the per-vehicle results
by evaluating two different properties of the predicted scene. Intention-only methods are excluded because they do not generate trajectory outputs, while strictly target-centered trajectory models are excluded because they predict one designated vehicle at a time and do not provide synchronized futures for the remaining vehicles in the scene. The retained baselines can generate trajectory predictions for multiple vehicles within the same local scene and are therefore evaluated using the same synchronized scene composition, prediction horizon, and scene-level metrics.

As shown in Table~\ref{tab:scene_level_metrics}, DSiGAT achieves the best
IACR and JDE results on both NGSIM I-80 and US-101. On I-80, the proposed
model obtains an IACR of 0.087 and a JDE of 1.22, compared with 0.109 and
1.41, respectively, for the strongest competing baseline, GRIP++. These
results correspond to a 20.18\% reduction in inter-agent collision rate and
a 13.48\% reduction in joint displacement error. A similar trend is observed
on US-101, where DSiGAT achieves an IACR of 0.093 and a JDE of 1.34,
compared with 0.116 and 1.53 for GRIP++. This represents reductions of
19.83\% in IACR and 12.42\% in JDE.

A clear pattern is also observed across the comparison methods. Social Conv
and MATF GAN produce weaker scene-level results, indicating that incorporating
surrounding-agent information does not necessarily guarantee that the
resulting multi-vehicle trajectories remain mutually compatible. GRIP,
GRIP++, and Dual Transformer improve scene-level performance through stronger
interaction modeling, but DSiGAT remains more effective on both datasets.
This advantage is consistent with the design of the proposed framework. The
dynamic scene graph represents evolving inter-vehicle relationships explicitly,
while relational edge features and temporal graph attention message passing
allow the model to capture how these interactions change throughout the
observation interval. Intention-guided trajectory decoding and scene-level
consistency learning further align maneuver reasoning with future motion,
thereby reducing physically conflicting predictions and better preserving the
relative configuration of the predicted traffic scene.

\begin{table*}[!htbp]
\centering
\caption{Scene-level prediction performance on the NGSIM I-80 and
US-101 test sets. Lower IACR indicates fewer physically conflicting
multi-vehicle predictions, while lower JDE indicates better preservation
of the relative spatial configuration of vehicles. Best results are shown
in \textcolor{bestred}{\textbf{red}}, and second-best results are
underlined.}
\label{tab:scene_level_metrics}
\renewcommand{\arraystretch}{1.08}
\setlength{\tabcolsep}{8pt}

\begin{tabular}{l|cc|cc}
\toprule
\multirow{2}{*}{\textbf{Method}}
& \multicolumn{2}{c|}{\textbf{NGSIM I-80}}
& \multicolumn{2}{c}{\textbf{NGSIM US-101}} \\
\cline{2-5}
& \textbf{IACR}$\downarrow$
& \textbf{JDE}$\downarrow$
& \textbf{IACR}$\downarrow$
& \textbf{JDE}$\downarrow$ \\
\midrule

Social Conv
& 0.164 & 1.92
& 0.171 & 2.06 \\

MATF GAN
& 0.151 & 1.84
& 0.158 & 1.97 \\

GRIP
& 0.126 & 1.58
& 0.133 & 1.71 \\

GRIP++
& \underline{0.109}
& \underline{1.41}
& \underline{0.116}
& \underline{1.53} \\

Dual Transformer
& 0.118 & 1.49
& 0.124 & 1.61 \\

\midrule

\rowcolor{red!5}
\textcolor{bestred}{\textbf{DSiGAT (Ours)}}
& \textcolor{bestred}{\textbf{0.087}}
& \textcolor{bestred}{\textbf{1.22}}
& \textcolor{bestred}{\textbf{0.093}}
& \textcolor{bestred}{\textbf{1.34}} \\

\bottomrule
\end{tabular}
\end{table*}

\subsection{Sensitivity and Robustness Analysis}
\subsubsection{Robustness to Input Noise}
To evaluate robustness to imperfect sensing and trajectory estimation,
zero-mean Gaussian noise is applied only to the continuous kinematic and
geometric input features at test time. These features include longitudinal
speed, longitudinal and lateral acceleration, longitudinal and lateral
displacement, pairwise longitudinal and lateral gaps, and relative
longitudinal and lateral speeds. The perturbation is defined as
\begin{equation}
\widetilde{x}=x+\eta,
\qquad
\eta\sim\mathcal{N}(0,\sigma^2),
\end{equation}
where $\sigma$ controls the perturbation magnitude.

Categorical and binary variables, including one-hot lane identity,
lane-relation encodings, validity masks, and graph connectivity, are not
perturbed. Time-to-collision values are recomputed from the perturbed
continuous gaps and relative speeds rather than being independently
corrupted. Noise is added after conversion to SI units and before feature
standardization, after which the original training-set normalization
statistics are applied. The evaluated noise levels are
$\sigma\in\{0.00,0.05,0.10,0.20,0.30,0.50\}$.

Table~\ref{tab:noise_robustness} reports the intention prediction
performance of DSiGAT under different Gaussian noise levels. As expected,
performance decreases as the standard deviation $\sigma$ increases. Overall
accuracy decreases from 90.12\% in the noise-free setting to 88.60\%,
86.52\%, 82.60\%, 77.34\%, and 68.04\% for
$\sigma=0.05$, $0.10$, $0.20$, $0.30$, and $0.50$, respectively. Even with this degradation, the results remain meaningful under mild and moderate perturbations. Under $\sigma=0.05$, the model retains F1-scores of
92.34\%, 88.12\%, and 85.34\% for LK, LLC, and RLC,
respectively. Under $\sigma=0.10$, the corresponding values remain at 91.12\%, 85.78\%, and 82.67\%. This suggests that the proposed scene-level interaction modeling is reasonably stable when the observed vehicle states contain small disturbances. The performance drop becomes more pronounced at higher noise levels, which is expected because stronger perturbations directly affect the motion and interaction cues used for maneuver reasoning.

A clear difference is also observed across the maneuver classes. Lane-keeping remains the most robust class under noise, while LLC and RLC degrade faster as perturbation increases. This is reasonable because lane-change prediction depends more heavily on subtle lateral motion and inter-vehicle interaction cues, which are more easily distorted by noisy observations than stable lane-keeping behavior. Still, the gradual degradation pattern indicates that DSiGAT does not fail abruptly under perturbation, but instead preserves useful predictive capability as the input quality deteriorates.

\begin{table*}[!htbp]
\centering
\caption{Robustness of DSiGAT to input Gaussian noise on NGSIM I-80 (unit: \%).
$\mathcal{N}(0,\sigma^2)$ denotes additive Gaussian noise applied only
to continuous kinematic and geometric features at test time. Categorical
lane encodings, masks, and graph connectivity are not perturbed.}
\label{tab:noise_robustness}
\scriptsize
\setlength{\tabcolsep}{4.5pt}
\renewcommand{\arraystretch}{1.15}
\begin{adjustbox}{max width=\textwidth}
\begin{tabular}{l c ccc ccc ccc}
\toprule

\multirow{2}{*}{\textbf{Input Noise}}
& \multirow{2}{*}{\textbf{Accuracy}}
& \multicolumn{3}{c}{\textbf{LK}}
& \multicolumn{3}{c}{\textbf{LLC}}
& \multicolumn{3}{c}{\textbf{RLC}} \\

\cmidrule(lr){3-5}\cmidrule(lr){6-8}\cmidrule(lr){9-11}

& & P & R & F1 & P & R & F1 & P & R & F1 \\

\midrule

$\sigma=0.00$
& \textbf{90.12}
& \textbf{92.25} & \textbf{94.68} & \textbf{93.45}
& \textbf{87.37} & \textbf{92.09} & \textbf{89.67}
& \textbf{84.43} & \textbf{90.22} & \textbf{87.23} \\

$\sigma=0.05$
& 88.60
& 90.84 & 93.89 & 92.34
& 85.72 & 90.65 & 88.12
& 82.44 & 88.45 & 85.34 \\

$\sigma=0.10$
& 86.52
& 89.32 & 92.99 & 91.12
& 83.08 & 88.66 & 85.78
& 79.47 & 86.13 & 82.67 \\

$\sigma=0.20$
& 82.60
& 86.35 & 90.65 & 88.45
& 78.23 & 84.46 & 81.23
& 74.62 & 81.96 & 78.12 \\

$\sigma=0.30$
& 77.34
& 81.83 & 86.77 & 84.23
& 72.15 & 79.06 & 75.45
& 68.54 & 76.58 & 72.34 \\

$\sigma=0.50$
& 68.04
& 73.64 & 79.24 & 76.34
& 61.74 & 69.38 & 65.34
& 58.35 & 67.16 & 62.45 \\

\bottomrule
\end{tabular}
\end{adjustbox}
\end{table*}

\subsection{Loss-Weight Sensitivity Analysis}

To examine the stability of the proposed framework with respect to its multi-objective training design, a loss-weight sensitivity analysis is conducted by varying the coefficients of the intention classification loss, trajectory regression loss, intention-guided consistency loss, and scene-level consistency loss.  Similar multi-task and scene-consistent prediction settings often require a reasonable trade-off between classification quality, trajectory accuracy, and scene compatibility, making sensitivity analysis a useful complement to the main experimental results.

\begin{figure*}
    \centering
    \includegraphics[width=1\linewidth]{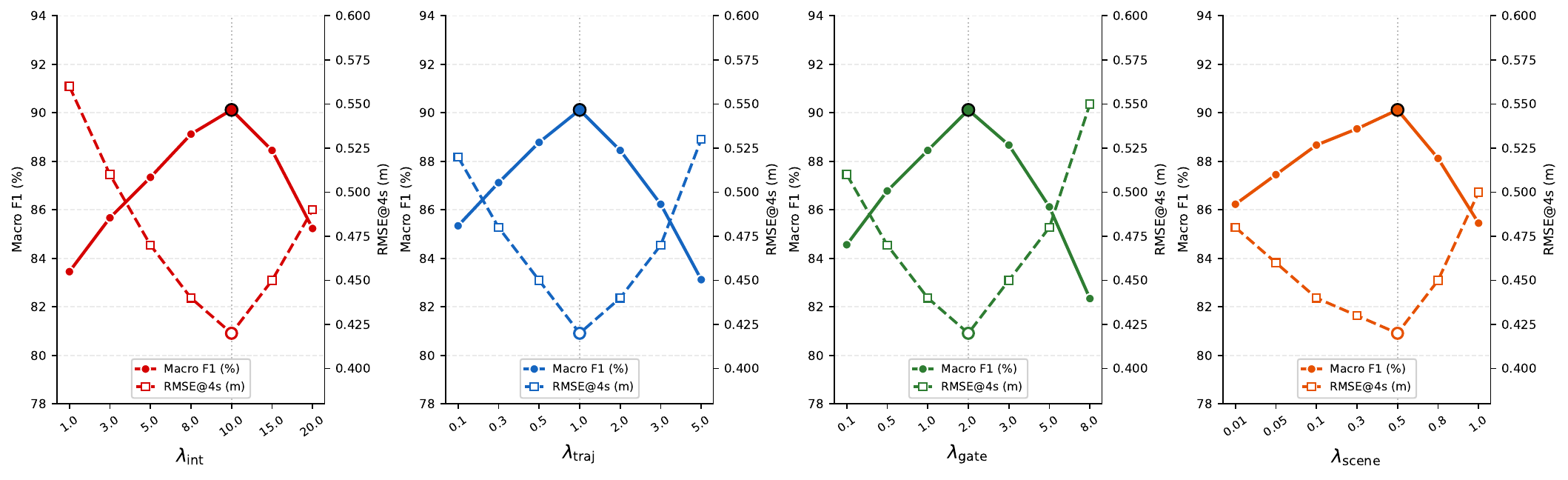}
    \caption{Sensitivity analysis of the loss weights in DSiGAT. The effects of $\lambda_{\mathrm{int}}$, $\lambda_{\mathrm{traj}}$, $\lambda_{\mathrm{gate}}$, and $\lambda_{\mathrm{scene}}$ are evaluated using Macro F1-score and RMSE@4s.}   
    \label{fig:loss_sensitivity}
\end{figure*}

Fig.~\ref{fig:loss_sensitivity} summarizes the influence of the four loss weights on Macro F1-score and RMSE@4s. A consistent pattern is observed across all four cases: the performance first improves as the corresponding weight increases, reaches a favorable operating point, and then degrades when the weight becomes too small or too large. This shows that each objective contributes meaningfully to the final model, but also that excessive emphasis on any single term weakens the balance between intention recognition and motion forecasting. 

For the intention loss, the best trade-off is obtained around $\lambda_{\mathrm{int}}=10$, where the Macro F1-score reaches its peak while RMSE@4s is near its minimum. When $\lambda_{\mathrm{int}}$ is set too low, the model does not learn sufficiently discriminative maneuver cues. When it is increased beyond the optimal range, the intention objective begins to dominate training, and the trajectory-related benefit becomes weaker. A similar behavior is observed for the trajectory loss, where $\lambda_{\mathrm{traj}}=1.0$ gives the best balance between class recognition and motion accuracy. Smaller values under-emphasize trajectory supervision, while larger values gradually reduce Macro F1 and increase RMSE@4s. 

The same trend is also evident for the consistency-related terms. The best result for the intention-guided consistency loss is obtained at $\lambda_{\mathrm{gate}}=2.0$, and the best result for the scene-level consistency loss is obtained around $\lambda_{\mathrm{scene}}=0.5$. In both cases, moderate weighting improves Macro F1 and reduces RMSE@4s, which suggests that these terms help align maneuver reasoning with future motion and encourage more coherent scene-level prediction. However, when either weight becomes too large, the optimization becomes over-constrained and both metrics begin to deteriorate. 
Based on these observations, $\lambda_{\mathrm{int}}=10$, $\lambda_{\mathrm{traj}}=1.0$, $\lambda_{\mathrm{gate}}=2.0$, and $\lambda_{\mathrm{scene}}=0.5$ are adopted in the final model. 

\subsection{Computational Efficiency}
\label{subsec:computational_efficiency}

Table~\ref{tab:efficiency_main} compares the computational efficiency of
DSiGAT with the main neural baselines. All models are evaluated on the same
NVIDIA A100 GPU using a batch size of one scene. Each input scene contains
up to 6 vehicles and 30 observation frames. Inference is performed in
evaluation mode with gradient computation disabled. Data loading, scene
construction, and preprocessing are excluded from the timing measurements.

Before measurement, 50 warm-up forward passes are performed to stabilize
GPU execution. The reported inference time is then averaged over 500
forward passes, with CUDA synchronization applied immediately before and
after each timed run. Peak GPU memory is measured during the forward pass
after resetting the CUDA memory statistics. Parameter counts include all
trainable model parameters, while GFLOPs are estimated for one complete
scene-level forward pass. All models are evaluated using single-precision
floating-point arithmetic under the same software and hardware environment.

As shown in Table~\ref{tab:efficiency_main}, DSiGAT contains 1.02 million
trainable parameters, which is substantially lower than VWC, Dual
Transformer, and MCLG and only moderately higher than the lightweight LSTM
baseline. The proposed framework records the lowest peak GPU memory usage among
the compared models and maintains moderate computational complexity at
1.29 GFLOPs per scene-level forward pass. These results indicate that the dynamic graph representation and
multi-vehicle prediction design do not require an excessively large model.

DSiGAT requires 167.68 ms to process one scene. Although this is slower
than the simpler target-centered CNN--LSTM and LSTM models, it is faster
than VWC, Dual Transformer, and MCLG. Relative to these stronger baselines,
DSiGAT reduces inference time by 28.51\%, 46.39\%, and 15.51\%,
respectively. The additional latency compared with the recurrent baselines
is attributable to constructing and processing the temporal interaction
graphs over multiple vehicles and observation frames. Nevertheless, the
results demonstrate that DSiGAT achieves scene-level joint intention and
trajectory prediction with moderate computational requirements and a
favorable balance between predictive performance and implementation cost.

\begin{table*}[!t]
\centering
\caption{Computational-efficiency comparison measured on an NVIDIA A100
GPU using a batch size of one scene. Inference time is averaged over 500
CUDA-synchronized forward passes after 50 warm-up runs. Best
results are shown in \textcolor{bestred}{\textbf{red}}, and second-best
results are underlined.}
\label{tab:efficiency_main}
\renewcommand{\arraystretch}{1.08}
\setlength{\tabcolsep}{8pt}

\begin{tabular}{lcccc}
\toprule
\textbf{Method}
& \textbf{Parameters}
& \textbf{Inference time}
& \textbf{Peak GPU memory}
& \textbf{GFLOPs} \\
& & \textbf{(ms/scene)} & \textbf{(MB)} & \\
\midrule

CNN--LSTM
& 1.24M
& \underline{12.34}
& 87.4
& \underline{0.34} \\

LSTM
& \textcolor{bestred}{\textbf{0.87M}}
& \textcolor{bestred}{\textbf{8.45}}
& \underline{43.2}
& \textcolor{bestred}{\textbf{0.18}} \\

VWC
& 2.13M
& 234.56
& 312.4
& 1.87 \\

Dual Transformer
& 3.45M
& 312.78
& 489.6
& 2.34 \\

MCLG
& 1.87M
& 198.45
& 267.8
& 1.45 \\

\midrule

\textbf{DSiGAT (Ours)}
& \underline{1.02M}
& 167.68
& \textcolor{bestred}{\textbf{20.2}}
& 1.29 \\

\bottomrule
\end{tabular}
\end{table*}

\section{Conclusion}
\label{conclusion}

This study presented DSiGAT, a scene-level dynamic graph framework for joint lane-change intention and trajectory prediction of multiple interacting vehicles. By modeling each local traffic scene as a dynamic interaction graph and learning future behavior through explicit multi-vehicle reasoning, the proposed framework moves beyond target-centered prediction and enables unified reasoning over maneuver intention, future motion, and scene-level compatibility.

The experimental results showed clear and consistent gains across datasets and evaluation settings. For overall intention prediction, DSiGAT achieved accuracies of 90.12\% on NGSIM I-80 and 90.97\% on US-101, with strong class-wise F1-scores across lane keeping, left lane change, and right lane change. The proposed framework also maintained strong early anticipation performance at $T=3$\,s before maneuver onset. For trajectory prediction, DSiGAT achieved the lowest RMSE at all evaluated horizons from 1 s to 4 s on both NGSIM datasets. Relative to GRIP++, the RMSE reductions ranged from 37.38\% to 50.00\% on I-80 and from 33.18\% to 38.95\% on US-101. On highD, DSiGAT achieved the best results at 1 s, 3 s, and 4 s and remained within 0.01 m of the best baseline at 2 s.

The added analyses further strengthened these findings. Scene-level evaluation showed that DSiGAT achieved the best inter-agent collision rate and joint displacement error on both NGSIM sites, indicating that the model improves not only per-vehicle prediction but also the joint coherence of the full traffic scene. The ablation study showed that the mean F1-score increased from 61.77\% for the baseline to 90.54\% for the full model, confirming the contribution of each component. Sensitivity and robustness analyses showed that the proposed framework remains stable under reasonable loss-weight settings and degrades gradually under increasing Gaussian noise. In addition, DSiGAT achieved these gains with only 1.02M parameters, 20.2 MB GPU memory, and 1.29 GFLOPs, showing that the framework remains computationally efficient despite its scene-level design.

\section{Declaration of Competing Interest}
The authors declare that they have no known competing financial interests or personal relationships that could have appeared to influence the work reported in this paper.

\section{Acknowledgment}
This work was supported by the National Science Foundation (NSF) Engines: Farms under Grant No. FAR0038157 and by the Upper Great Plains Transportation Institute (UGPTI) at North Dakota State University (USA) through the USDOT University Transportation Center (UTC) program, Project No. CTIPS-51

\printcredits

\bibliographystyle{cas-model2-names}

\bibliography{cas-refs}



\end{document}